\documentclass{article}

\usepackage{microtype}
\usepackage{graphicx}
\usepackage{subcaption}
\usepackage{booktabs} 

\usepackage{hyperref}

\usepackage[preprint]{icml2026}

\usepackage{amsmath}
\usepackage{amssymb}
\usepackage{mathtools}
\usepackage{amsthm}

\usepackage[capitalize,noabbrev]{cleveref}

\theoremstyle{plain}

\theoremstyle{definition}

\theoremstyle{remark}

\usepackage[textsize=tiny]{todonotes}

\usepackage{tcolorbox}
\usepackage{xspace}
\usepackage{amsmath}
\usepackage{tabularx}
\usepackage{multirow}
\usepackage{makecell}
\usepackage{booktabs}
\usepackage{enumitem}
\usepackage{tcolorbox}
\usepackage{tikz}
\usepackage{pgfplots}
\pgfplotsset{compat=1.18}
\usepackage{pifont}
\usepackage{colortbl}

\usepackage{svg}
\usepackage[framemethod=TikZ]{mdframed}
\usepgfplotslibrary{groupplots}
\usetikzlibrary{patterns}

\icmltitlerunning{APR: Penalizing Structural Redundancy in Large Reasoning Models via Anchor-based Process Rewards}

\begin{document}

\twocolumn[
  \icmltitle{APR: Penalizing Structural Redundancy in Large Reasoning Models \\ via Anchor-based Process Rewards}



  \icmlsetsymbol{equal}{*}

  \begin{icmlauthorlist}
    \icmlauthor{Kaiyan Chang}{sch}
    \icmlauthor{Chenwei Zhu}{sch}
    \icmlauthor{Yingfeng Luo}{sch}
    \icmlauthor{Yifu Huo}{sch}
    \icmlauthor{Chenglong Wang}{sch}
    \icmlauthor{Xiaoqian Liu}{sch}
    \icmlauthor{Qiaozhi He}{sch}
    \icmlauthor{Tong Xiao}{sch,comp}
    \icmlauthor{Zhengtao Yu}{sch2}
    \icmlauthor{Jingbo Zhu}{sch,comp}
  \end{icmlauthorlist}

  \icmlaffiliation{comp}{NiuTrans Research}
  \icmlaffiliation{sch}{Northeastern University, China}
  \icmlaffiliation{sch2}{Kunming University of Science and Technology}

  \icmlcorrespondingauthor{Tong Xiao}{xiaotong@mail.neu.edu.cn}

  \icmlkeywords{Large Reasoning Models, Efficient Reasoning, Process Reward, Reinforcement Learning}

  \vskip 0.3in
]



\printAffiliationsAndNotice{}  

\begin{abstract}
Test-Time Scaling (TTS) has significantly enhanced the capabilities of Large Reasoning Models (LRMs) but introduces a critical side-effect known as Overthinking.
We conduct a preliminary study to rethink this phenomenon from a fine-grained perspective.
We observe that LRMs frequently conduct repetitive self-verification without revision even after obtaining the final answer during the reasoning process. 
We formally define this specific position where the answer first stabilizes as the Reasoning Anchor.
By analyzing pre- and post-anchor reasoning behaviors, we uncover the structural redundancy fixed in LRMs: the meaningless repetitive verification after deriving the first complete answer, which we term the Answer-Stable Tail (AST).
Motivated by this observation, we propose Anchor-based Process Reward (APR), a structure-aware reward shaping method that localizes the reasoning anchor and penalizes exclusively the post-anchor AST.
Leveraging the policy optimization algorithm suitable for length penalties, our APR models achieved the performance-efficiency Pareto frontier at 1.5B and 7B scales averaged across five mathematical reasoning datasets while requiring substantially fewer computational resources for RL training.
\end{abstract}

\section{Introduction}
In recent years, Large Reasoning Models (LRMs)~\cite{openai_o1_system_card, team2024qwq, guo2025deepseek} have achieved significant performance breakthroughs on complex tasks by Test-Time Scaling (TTS)~\cite{snell2024scaling}. 
However, reasoning length extension has triggered a widespread phenomenon called \textit{Overthinking}~\cite{chen2024not, luo2025o1}, where the model generates substantial redundant reasoning behaviors that make no contribution to the final answer~\cite{su2025between}.
The resulting prohibitive computational costs and impractical response latency consequently hinder the scalability of LRMs in real-world serving systems.
Therefore, it is urgent for LRMs to mitigate reasoning redundancy to enhance efficiency without sacrificing performance.

The Reinforcement Learning with Verifiable Rewards (RLVR)~\cite{lambert2024tulu, shao2024deepseekmath} paradigm has not only successfully established LRMs with superior reasoning capabilities but is also increasingly leveraged to investigate efficiency optimization.
Existing research primarily integrates a length penalty into the reward function. 
One category imposes hard constraints, directly truncating or penalizing outputs exceeding the optimal response-level length~\cite{aggarwal2025l1, hou2025thinkprune}; the other attempts difficulty-adaptive adjustments, dynamically tuning target length based on outcome-level accuracy~\cite{luo2025o1, xiang2025just}.
However, we characterize both response-level and outcome-level constraints as coarse-grained global penalties. 
This holistic perspective ignores the intrinsic variations in \textit{Information Contribution} across different reasoning phases, posing a risk that the model will discard necessary reasoning steps when compressing length.

\begin{figure}
    \centering
    \includegraphics[width=0.99\linewidth]{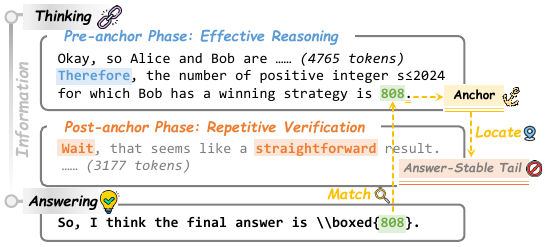}
    \caption{Schematic diagram of structural redundancy in LRMs. The reasoning anchor splits the trace into an information-dense pre-anchor phase of effective derivation, correction and first conclusion, and an information-sparse post-anchor phase of repetitive self-verification without revision, termed the \emph{Answer-Stable Tail (AST)}.}
    \label{fig:redundancy}
    \vspace{-0.5em}
\end{figure}

In this paper, we aim to analyze the sources of redundancy in LRM reasoning traces at a finer granularity.
Motivated by our observation that LRMs rarely revise the initial answer even in erroneous reasoning traces, we posit that overthinking should not be constrained solely to redundancy following a \emph{correct} answer.
We therefore introduce a reasoning anchor to identify the precise position where the final answer first emerges in the reasoning trace.
Extensive empirical analyses reveal a common phenomenon: regardless of whether the response is incorrect or truncated, the model typically enters a verification loop immediately after deriving its first complete answer.
Notably, this initial answer remains stable throughout the subsequent generation, contributing no further substantive information.
From an information-gain perspective, we rethink overthinking as an intrinsic structural redundancy in LRMs. As illustrated in Fig.~\ref{fig:redundancy}, the pre-anchor phase is responsible for effective derivation, correction and initial answer generation, exhibiting high information density. In contrast, the post-anchor phase is dominated by repetitive verification of the existing answer without revision, exhibiting sparse information gain.
We term this redundant suffix the \emph{Answer-Stable Tail (AST)}, a critical redundant structure that should be reduced to improve LRM efficiency without degrading performance.

Inspired by this insight, we propose \emph{Anchor-based Process Reward (APR)}, a structure-aware reward shaping mechanism specifically designed to mitigate the inefficiency introduced by AST.
Unlike prior global length penalties, we develop reliable anchor localization methods (rule-based and model-based) to identify the AST and penalize exclusively this redundant segment.
Leveraging the Direct Alignment Policy Optimization algorithm~\cite{yu2025dapo}, we achieve the Pareto frontier of average reasoning performance and efficiency across five mathematical reasoning datasets on two representative LRMs, benchmarking against six open-source efficient reasoning models. 
Specifically, our APR-1.5B achieves a 52.8\% reduction in generation length with a 16.3\% accuracy improvement, while APR-7B shortens reasoning traces by 56.7\% accompanied by an 11.5\% accuracy increase.
Remarkably, our method consumes fewer computational resources during Reinforcement Learning (RL) than comparable methods, highlighting that dense and accurate process reward signals are pivotal for training efficiency.
The contributions of our work are three-fold:
\begin{itemize}[leftmargin=*]
    \item \textbf{New Perspective on Redundancy}: We rethink overthinking through the lens of information gain, identifying the \emph{Answer-Stable Tail (AST)} as a distinct form of structural redundancy regardless of answer correctness.
    \item \textbf{Structure-Aware Reward Mechanism}: We propose \emph{Anchor-based Process Reward (APR)}, a fine-grained optimization method that leverages anchor localization to precisely eliminate the inefficient AST. 
    \item \textbf{Superior Training Efficiency}: We validate that dense and accurate process reward signals can improve RL training efficiency, achieving favorable accuracy-efficiency trade-offs with comparatively fewer training resources.
\end{itemize}

\section{Related Work}
\subsection{Overthinking in LLM Reasoning}
Test-time Scaling (TTS)~\cite{snell2024scaling} extends the intelligence boundaries of Large Language Models (LLMs) through increased inference-time compute, which marks a shift in LLM's thinking patterns from ``System 1'' to ``System 2'' similar to human cognition~\cite{kaneman2011thinking}.
In practice, TTS is broadly categorized into training-free~\cite{zhang2025llama, chang2025step} and training-based~\cite{shao2024deepseekmath, muennighoff2025s1} strategies.
While both approaches encourage models to explore longer Chain-of-Thought (CoT) for performance gains, the training-based RLVR paradigm has been proven to decrease the model's output entropy~\cite{yue2025does}.
In other words, the model's thinking patterns become less diverse, leading it to generate lengthy reasoning even for simple tasks, a phenomenon known as overthinking~\cite{chen2024not}.

To this end, existing works design a learned router to automatically determine the optimal reasoning strategy based on the difficulty of problems, such as assigning simple problems to smaller models (System 1) and complex ones to larger models (System 2)~\cite{liang2025thinkswitcher, he2025self, openai2025gpt5}.
However, these methods rely on the accuracy of external modules and increase deployment complexity, failing to address the model's intrinsic tendency to generate redundancy.

\subsection{Efficient Reasoning via Reinforcement Learning}
In practice, an ideal LRM should adaptively switch between System 1 and System 2, so the focus of efficient reasoning remains centered on intrinsic model optimization. 
Given that RLVR has become the dominant post-training paradigm for LRMs, recent efforts further explores the RL objective by integrating response length into the reward function.

One line of work directly penalizes verbose response based on the target length, whose definition varies across implementations. 
Strategies include normalizing length within the group~\cite{team2025kimi, arora2025training, cheng2025optimizing},  setting a fixed truncation length~\cite{aggarwal2025l1, liu2025dler, li2025leash} or even dynamically updated computational budget~\cite{hammoud2025train}, and leveraging LLMs to evaluate conciseness~\cite{dumitru2025conciserl, wang2025efficient}.

To mitigate the performance risks of indiscriminate length penalties, another line of work couples outcome accuracy with reasoning length, aiming to adaptively adjust reasoning depth based on estimated problem difficulty. 
ShorterBetter~\cite{yi2025shorterbetter} selects the shortest correct sample among rollouts as the optimal target; A-DLP~\cite{su2025thinking} derives the penalty coefficient from the accuracy gap relative to a reference model; ALP~\cite{xiang2025just} and Laser~\cite{liu2025learnreasonefficientlyadaptive} use empirical accuracy within the group to indicate difficulty.

However, these methods operate as coarse-grained global optimizations that fail to distinguish specific redundant segments, thereby risking the removal of necessary reasoning steps, ultimately degrading performance.
Therefore, we focus on analyzing redundancy at a finer granularity, aiming to provide precise feedback signals on length redundancy during the RL process.

\section{Preliminary Study}\label{sec:preliminary}
In this section, we conduct a fine-grained empirical analysis on the overthinking phenomenon of LRMs.
We start by defining the reasoning anchor to quantify redundancy.
Subsequently, we introduce effective methods to precisely locate these anchors within complex context.
Finally, through a deep dive into the post-anchor reasoning behavior, we formally identify the \emph{Answer-Stable Tail (AST)} as the redundant suffix segment spanning from the first stable appearance of the final answer to the end of the thinking process.

\subsection{Anchor Definition}
\label{sec:anchor_def}
Distinct from prior findings~\cite{chen2024not, luo2025o1} that characterize redundancy in overthinking as the sequence generated after the first occurrence of the \emph{correct} answer, we observe that even when the model’s derived answer is incorrect, the reasoning trace often persists in a repetitive self-verification loop without any revision. We identify this indiscriminate repetition of the established answer as \emph{structural redundancy}, defined as the sequence generated after the answer first becomes stable, where stability is assessed with respect to a reference answer $y_{\text{ref}}$.

Accordingly, we define the \emph{reasoning anchor} as the structural boundary marking the emergence of this redundancy.
Specifically, we segment the thinking process into a sequence of sentences
$S=(s_1,\dots,s_N)$, where $N$ is the number of sentences.
Let $\mathrm{end}(s_i)$ denote the global token index of the last token of sentence $s_i$.
We aim to locate the index $k^* \in \{1, \dots, N\}$ of the critical sentence in which the reference answer \textit{first} appears.
The anchor is then defined as the position of the last token of this sentence:
\begin{equation}
\label{eq:anchor_func}
t_{\text{anc}}(y, y_{\text{ref}}) = \mathrm{end}(s_{k^*})
\end{equation}

Despite its conceptual simplicity, accurately locating $t_{\text{anc}}$ poses non-trivial challenges:
(1) Unlike the structured final answer, the thinking process lacks standardized delimiters (e.g., \texttt{\textbackslash boxed\{\}}), rendering regex-based extraction of intermediate answers unreliable.
(2) Linguistic patterns for the conclusion phase are highly diverse, where common connectors (e.g., \texttt{so}) often trigger false positives by indicating intermediate steps rather than the final solution.

\subsection{Anchor Localization}\label{3.2}
To address these challenges, we explore two localization strategies to identify the index of the target sentence $k^*$: Rule-based and Model-based.

\paragraph{Rule-based Localization.}
Rule-based localization relies on the observation that a valid anchor resides within a conclusion sentence followed by a verification sentence.
We identify $k^*$ by narrowing down candidate sentences using two screening criteria.

First, let $\mathcal{A}(\cdot)$ be an extraction function mapping a sentence to a mathematical expression, and $\mathcal{E}(\cdot,\cdot)\in\{0,1\}$ be an equivalence function testing whether two answers are mathematically equivalent. The set of sentences matching the reference answer is:
\begin{equation}
\mathcal{I}_{\text{math}} = \left\{ i \in \{1,\dots,N\} \;\middle|\; \mathcal{E}\big(\mathcal{A}(s_i),\, y_{\text{ref}}\big)=1 \right\}
\end{equation}

Second, to filter out false positives such as coincidental occurrences of the answer value in the problem statement or intermediate reasoning steps, we apply context constraints. Let $\text{Con}(\cdot)$ and $\text{Ver}(\cdot)$ denote the presence of conclusion indicators and verification patterns (see Appendix~\ref{app:keywords} for the detailed keyword lists), respectively. The set of contextually valid sentences is:
\begin{equation}
\mathcal{I}_{\text{ctx}} = \left\{ i \in \{1,\dots,N\} \;\middle|\; \text{Con}(s_i) \lor \text{Ver}(s_{i+1}) \right\}
\end{equation}

We determine $k^*$ as the index of the earliest sentence satisfying both conditions (the strict intersection):
\begin{equation}
\label{eq:rule_based_min}
k^* = \min \Big( (\mathcal{I}_{\text{math}} \cap \mathcal{I}_{\text{ctx}}) \cup \{N\} \Big)
\end{equation}
If the intersection is empty, the default $k^*=N$ implies the anchor is set to the end of the thinking process, resulting in zero redundancy.

\paragraph{Model-based Localization.}
Given the inherent rigidity of rule-based methods in handling complex formats (e.g., intervals, sets) and distractors, we introduce a Model-based Anchor Locator to explore a more flexible, semantic-aware alternative.
We model the identification of $k^*$ as a sentence extraction task.
Specifically, we constructed a high-quality dataset of 12k samples annotated by Gemini3-Flash and fine-tuned a Qwen3-8B model~\cite{yang2025qwen3} (see Appendix~\ref{app:locator} for implementation details).
Taking the thinking process and final answer as input, the locator is trained to directly predict the sentence $s_{k^*}$ that contains the first derived reference answer.
Evaluation on a manually verified test set ($n=500$) shows an exact match accuracy of 66.4\% while the extraction validity is 100\% indicating no hallucinated content.
Further error analysis reveals that 80.1\% of mismatches exhibit only a minor offset (e.g., pointing to a sentence slightly earlier or later than the ground truth), which we consider acceptable for redundancy identification.

\definecolor{qred}{HTML}{F57C6E}
\definecolor{qgreen}{HTML}{84C3B7}
\definecolor{qblue}{HTML}{71B8ED}
\definecolor{qpink}{HTML}{F2A8DA}
\tikzset{
  solidbar/.style={},
  hatchbar/.style={
    postaction={draw=none, pattern=north east lines}
  }
}
\begin{figure*}[ht]
\centering
\begin{minipage}{\textwidth}
\centering
\small
\begin{tikzpicture}
\def\startx{0}      
\def\spacing{2.5}   
\def\boxwidth{0.5}  
\def\boxheight{0.3} 

\definecolor{lightgray}{RGB}{230,230,230}
\definecolor{darkgray}{RGB}{150,150,150}

\foreach \type/\color/\pattern/\i in {
    0/qred/none/0,     
    1/qgreen/none/1,   
    2/qblue/none/2,    
    3/qpink/none/3,    
    4/black/none/4,    
    5/lightgray/darkgray/5  
} {
    \pgfmathsetmacro{\x}{\startx + \i*\spacing}
    
    \ifnum\i=5
        \draw[fill=lightgray, draw=black, line width=0.5pt, pattern=north east lines, pattern color=darkgray] 
            (\x,0) rectangle (\x+\boxwidth,\boxheight);
    \else
        \ifnum\i=4
            \draw[fill=white, draw=black, line width=0.5pt] 
                (\x,0) rectangle (\x+\boxwidth,\boxheight);
        \else
            \draw[fill=white, draw=\color, line width=0.5pt] 
                (\x,0) rectangle (\x+\boxwidth,\boxheight);
        \fi
    \fi
}

\node[anchor=west, font=\scriptsize, inner sep=2pt] at (\startx+0*\spacing+\boxwidth+0.1,\boxheight/2) {DS-1.5B};
\node[anchor=west, font=\scriptsize, inner sep=2pt] at (\startx+1*\spacing+\boxwidth+0.1,\boxheight/2) {DS-7B};
\node[anchor=west, font=\scriptsize, inner sep=2pt] at (\startx+2*\spacing+\boxwidth+0.1,\boxheight/2) {Qwen3-8B};
\node[anchor=west, font=\scriptsize, inner sep=2pt] at (\startx+3*\spacing+\boxwidth+0.1,\boxheight/2) {QwQ-32B};
\node[anchor=west, font=\scriptsize, inner sep=2pt] at (\startx+4*\spacing+\boxwidth+0.1,\boxheight/2) {Total Length};
\node[anchor=west, font=\scriptsize, inner sep=2pt] at (\startx+5*\spacing+\boxwidth+0.1,\boxheight/2) {Redundant Length};

\end{tikzpicture}
\end{minipage}

\vspace{2pt}
\begin{minipage}[t]{0.275\textwidth}
\centering
\begin{tikzpicture}
\begin{axis}[
    width=\linewidth,
    height=0.85\linewidth,
    ymajorgrids,
    grid style=dashed,
    ybar stacked,
    bar width=8pt,
    xtick={0,1,2,3,4,5},
    xticklabels={AIME24, AIME25,AMC, MATH500, Minerva, Olympiad},
    xticklabel style={rotate=90, anchor=east},
    xmin=-0.5,
    xmax=5.5,
    ymin=0,
    ymax=6000,
    ytick={0,2000,4000,6000},
    ticklabel style={font=\tiny},
    label style={font=\scriptsize},
    legend style={
        font=\tiny,
        fill opacity=0.5,
        draw=none,
        text opacity=1,
        at={(axis description cs:0.98,0.98)},
        anchor=north east
    },
    enlarge x limits=0.05
]

\draw[dashed, gray, thick] (axis cs:-1,1299) -- (axis cs:6,1299);
\node[anchor=west, font=\tiny, color=gray] at (axis cs:2.1,1299) [yshift=5pt, xshift=0.85em] {Avg.$\rho$ = 38.1\%};

\addplot[fill=qred!10,draw=qred,solidbar,hatchbar,pattern color=qred,    nodes near coords,  
    point meta=explicit symbolic,  
    every node near coord/.style={
    font=\tiny,
    color=black,
    anchor=south,
    fill=white,          
    fill opacity=0.7,    
    text opacity=1,      
    inner sep=1pt,       
    outer sep=0pt,       
    rounded corners=1pt, 
}] coordinates {
    (0,1383) 
    (1,1191) 
    (2,1531) 
    (3,1078) 
    (4,1089) 
    (5,1523)
};
\addplot[fill=white,draw=qred] coordinates {
    (0,2859) 
    (1,3049) 
    (2,1812) 
    (3,1302) 
    (4,2211) 
    (5,1962)
};

\legend{}
\end{axis}
\end{tikzpicture}

\end{minipage}%
\hspace{-2em}
\begin{minipage}[t]{0.275\textwidth}
\centering
\begin{tikzpicture}
\begin{axis}[
    width=\linewidth,
    height=0.85\linewidth,
    ymajorgrids,
    grid style=dashed,
    ybar stacked,
    bar width=8pt,
    xtick={0,1,2,3,4,5},
    xticklabels={AIME24, AIME25,AMC, MATH500, Minerva, Olympiad},
    xticklabel style={rotate=90, anchor=east},
    xmin=-0.5,
    xmax=5.5,
    ymin=0,
    ymax=6000,
    ytick={0,2000,4000,6000},
    ticklabel style={font=\tiny},
    label style={font=\scriptsize},
    legend style={
        font=\tiny,
        fill opacity=0.5,
        draw=none,
        text opacity=1,
        at={(axis description cs:0.98,0.98)},
        anchor=north east
    },
    enlarge x limits=0.05
]

\draw[dashed, gray, thick] (axis cs:-1,1156) -- (axis cs:6,1156);
\node[anchor=west, font=\tiny, color=gray] at (axis cs:2.1,1156) [yshift=5pt, xshift=0.85em] {Avg.$\rho$ = 36.1\%};

\addplot[fill=qgreen!10,draw=qgreen,solidbar,hatchbar,pattern color=qgreen] coordinates {
    (0,1126) (1,970) (2,1391) 
    (3,1075) (4,1021) (5,1354)
};
\addplot[fill=white,draw=qgreen] coordinates {
    (0,2955) (1,2881) (2,1890) 
    (3,1303) (4,1964) (5,1870)
};

\legend{}
\end{axis}
\end{tikzpicture}

\end{minipage}%
\hspace{-2em}
\begin{minipage}[t]{0.275\textwidth}
\centering
\begin{tikzpicture}
\begin{axis}[
    width=\linewidth,
    height=0.85\linewidth,
    ymajorgrids,
    grid style=dashed,
    ybar stacked,
    bar width=8pt,
    xtick={0,1,2,3,4,5},
    xticklabels={AIME24, AIME25,AMC, MATH500, Minerva, Olympiad},
    xticklabel style={rotate=90, anchor=east},
    xmin=-0.5,
    xmax=5.5,
    ymin=0,
    ymax=6000,
    ytick={0,2000,4000,6000},
    ticklabel style={font=\tiny},
    label style={font=\scriptsize},
    legend style={
        font=\tiny,
        fill opacity=0.5,
        draw=none,
        text opacity=1,
        at={(axis description cs:0.98,0.98)},
        anchor=north east
    },
    enlarge x limits=0.05
]

\draw[dashed, gray, thick] (axis cs:-1,2243) -- (axis cs:6,2243);
\node[anchor=west, font=\tiny, color=gray] at (axis cs:2.1,2243) [yshift=-5pt, xshift=0.85em] {Avg.$\rho$ = 54.3\%};

\addplot[fill=qblue!10,draw=qblue,solidbar,hatchbar,pattern color=qblue] coordinates {
    (0,3092) (1,2591) (2,2606) 
    (3,1706) (4,1337) (5,2127)
};
\addplot[fill=white,draw=qblue] coordinates {
    (0,2449) (1,1978) (2,1478) 
    (3,1340) (4,2127) (5,1733)
};

\legend{}
\end{axis}
\end{tikzpicture}

\end{minipage}%
\hspace{-2em}
\begin{minipage}[t]{0.275\textwidth}
\centering
\begin{tikzpicture}
\begin{axis}[
    width=\linewidth,
    height=0.85\linewidth,
    ymajorgrids,
    grid style=dashed,
    ybar stacked,
    bar width=8pt,
    xtick={0,1,2,3,4,5},
    xticklabels={AIME24, AIME25,AMC, MATH500, Minerva, Olympiad},
    xticklabel style={rotate=90, anchor=east},
    xmin=-0.5,
    xmax=5.5,
    ymin=0,
    ymax=6000,
    ytick={0,2000,4000,6000},
    ticklabel style={font=\tiny},
    label style={font=\scriptsize},
    legend style={
        font=\tiny,
        fill opacity=0.5,
        draw=none,
        text opacity=1,
        at={(axis description cs:0.98,0.98)},
        anchor=north east
    },
    enlarge x limits=0.05
]

\draw[dashed, gray, thick] (axis cs:-1,1914) -- (axis cs:6,1914);
\node[anchor=west, font=\tiny, color=gray] at (axis cs:2.1,1914) [yshift=5pt, xshift=0.85em] {Avg.$\rho$ = 48.7\%};

\addplot[fill=qpink!10,draw=qpink,solidbar,hatchbar,pattern color=qpink] coordinates {
    (0,1957) (1,2790) (2,2371) 
    (3,1321) (4,1157) (5,1890)
};
\addplot[fill=white,draw=qpink] coordinates {
    (0,2899) (1,2319) (2,1682) 
    (3,1295) (4,1889) (5,1883)
};

\legend{}
\end{axis}
\end{tikzpicture}

\end{minipage}%

\vspace{-0.4em}
{\small (a) Redundancy Ratios based on Rule-based Localization}
\vspace{0.4em}

\begin{minipage}[t]{0.275\textwidth}
\centering
\begin{tikzpicture}
\begin{axis}[
    width=\linewidth,
    height=0.85\linewidth,
    ymajorgrids,
    grid style=dashed,
    ybar stacked,
    bar width=8pt,
    xtick={0,1,2,3,4,5},
    xticklabels={AIME24, AIME25,AMC, MATH500, Minerva, Olympiad},
    xticklabel style={rotate=90, anchor=east},
    xmin=-0.5,
    xmax=5.5,
    ymin=0,
    ymax=6000,
    ytick={0,2000,4000,6000},
    ticklabel style={font=\tiny},
    label style={font=\scriptsize},
    legend style={
        font=\tiny,
        fill opacity=0.5,
        draw=none,
        text opacity=1,
        at={(axis description cs:0.98,0.98)},
        anchor=north east
    },
    enlarge x limits=0.05
]

\draw[dashed, gray, thick] (axis cs:-1,1285) -- (axis cs:6,1285);
\node[anchor=west, font=\tiny, color=gray] at (axis cs:2.1,1285) [yshift=5pt, xshift=0.85em] {Avg.$\rho$ = 38.5\%};

\addplot[fill=qred!10,draw=qred,solidbar,hatchbar,pattern color=qred] coordinates {
    (0,1094) (1,1013) (2,1300) 
    (3,1211) (4,1551) (5,1540)
};
\addplot[fill=white,draw=qred] coordinates {
    (0,3148) (1,3227) (2,2043) 
    (3,1169) (4,1749) (5,1945)
};

\legend{}
\end{axis}
\end{tikzpicture}

\end{minipage}%
\hspace{-2em}
\begin{minipage}[t]{0.275\textwidth}
\centering
\begin{tikzpicture}
\begin{axis}[
    width=\linewidth,
    height=0.85\linewidth,
    ymajorgrids,
    grid style=dashed,
    ybar stacked,
    bar width=8pt,
    xtick={0,1,2,3,4,5},
    xticklabels={AIME24, AIME25,AMC, MATH500, Minerva, Olympiad},
    xticklabel style={rotate=90, anchor=east},
    xmin=-0.5,
    xmax=5.5,
    ymin=0,
    ymax=6000,
    ytick={0,2000,4000,6000},
    ticklabel style={font=\tiny},
    label style={font=\scriptsize},
    legend style={
        font=\tiny,
        fill opacity=0.5,
        draw=none,
        text opacity=1,
        at={(axis description cs:0.98,0.98)},
        anchor=north east
    },
    enlarge x limits=0.05
]

\draw[dashed, gray, thick] (axis cs:-1,1190) -- (axis cs:6,1190);
\node[anchor=west, font=\tiny, color=gray] at (axis cs:2.1,1190) [yshift=5pt, xshift=0.85em] {Avg.$\rho$ = 37.8\%};

\addplot[fill=qgreen!10,draw=qgreen,solidbar,hatchbar,pattern color=qgreen] coordinates {
    (0,979) (1,940) (2,1125) 
    (3,1222) (4,1457) (5,1419)
};
\addplot[fill=white,draw=qgreen] coordinates {
    (0,3102) (1,2911) (2,2156) 
    (3,1156) (4,1528) (5,1805)
};

\legend{}
\end{axis}
\end{tikzpicture}

\end{minipage}%
\hspace{-2em}
\begin{minipage}[t]{0.275\textwidth}
\centering
\begin{tikzpicture}
\begin{axis}[
    width=\linewidth,
    height=0.85\linewidth,
    ymajorgrids,
    grid style=dashed,
    ybar stacked,
    bar width=8pt,
    xtick={0,1,2,3,4,5},
    xticklabels={AIME24, AIME25,AMC, MATH500, Minerva, Olympiad},
    xticklabel style={rotate=90, anchor=east},
    xmin=-0.5,
    xmax=5.5,
    ymin=0,
    ymax=6000,
    ytick={0,2000,4000,6000},
    ticklabel style={font=\tiny},
    label style={font=\scriptsize},
    legend style={
        font=\tiny,
        fill opacity=0.5,
        draw=none,
        text opacity=1,
        at={(axis description cs:0.98,0.98)},
        anchor=north east
    },
    enlarge x limits=0.05
]

\draw[dashed, gray, thick] (axis cs:-1,2276) -- (axis cs:6,2276);
\node[anchor=west, font=\tiny, color=gray] at (axis cs:2.1,2276) [yshift=5pt, xshift=0.85em] {Avg.$\rho$ = 56.3\%};

\addplot[fill=qblue!10,draw=qblue,solidbar,hatchbar,pattern color=qblue] coordinates {
    (0,2837) (1,2522) (2,2165) 
    (3,1956) (4,1981) (5,2196)
};
\addplot[fill=white,draw=qblue] coordinates {
    (0,2704) (1,2047) (2,1919) 
    (3,1090) (4,1483) (5,1664)
};

\legend{}
\end{axis}
\end{tikzpicture}

\end{minipage}%
\hspace{-2em}
\begin{minipage}[t]{0.275\textwidth}
\centering
\begin{tikzpicture}
\begin{axis}[
    width=\linewidth,
    height=0.85\linewidth,
    ymajorgrids,
    grid style=dashed,
    ybar stacked,
    bar width=8pt,
    xtick={0,1,2,3,4,5},
    xticklabels={AIME24, AIME25,AMC, MATH500, Minerva, Olympiad},
    xticklabel style={rotate=90, anchor=east},
    xmin=-0.5,
    xmax=5.5,
    ymin=0,
    ymax=6000,
    ytick={0,2000,4000,6000},
    ticklabel style={font=\tiny},
    label style={font=\scriptsize},
    legend style={
        font=\tiny,
        fill opacity=0.5,
        draw=none,
        text opacity=1,
        at={(axis description cs:0.98,0.98)},
        anchor=north east
    },
    enlarge x limits=0.05
]

\draw[dashed, gray, thick] (axis cs:-1,1816) -- (axis cs:6,1816);
\node[anchor=west, font=\tiny, color=gray] at (axis cs:2.1,1816) [yshift=5pt, xshift=0.85em] {Avg.$\rho$ = 47.5\%};

\addplot[fill=qpink!10,draw=qpink,solidbar,hatchbar,pattern color=qpink] coordinates {
    (0,1806) (1,2483) (2,1666) 
    (3,1462) (4,1605) (5,1875)
};
\addplot[fill=white,draw=qpink] coordinates {
    (0,3050) (1,2623) (2,2387) 
    (3,1154) (4,1441) (5,1898)
};

\legend{}
\end{axis}
\end{tikzpicture}

\end{minipage}%

\vspace{-0.4em}
{\small (b) Redundancy Ratios based on Model-based Localization}
\vspace{0.4em}

\caption{
\textbf{Redundancy Ratios ($\rho$)} across four LRMs on various mathematical reasoning datasets.
}
\label{fig:redundancy_1}
\end{figure*}
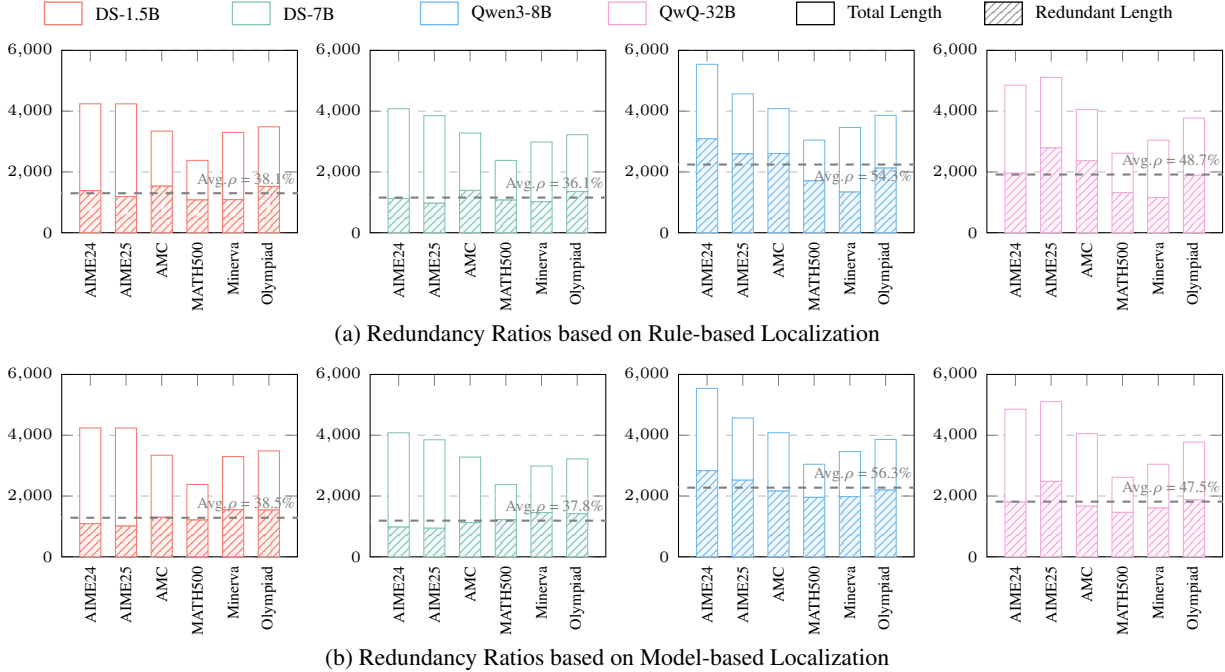
\subsection{Empirical Analysis of Structural Redundancy}
\label{sec:empirical_analysis}

To validate our hypothesis regarding structural redundancy, we conducted extensive empirical analyses using the two anchor localization methods on four base LRMs and six mathematical reasoning benchmarks.\footnote{LRMs: DeepSeek-R1-Distill-Qwen-1.5/7B (DS-1.5/7B), Qwen3-8B, and QwQ-32B\cite{qwq32b}. Datasets: AIME24, AIME25, AMC, MATH500, Minerva, and Olympiad Bench}.

We utilized the model's self-generated final answer $\hat{y}$ as the reference $y_{\text{ref}}$ to determine the reasoning anchor $t_{\text{anc}}(y,y_{\text{ref}})$. 
Formally, let $T_{\text{think}}$ denote the end position of the thinking process.
We define the redundancy length $L_{\text{red}}$ as:
\begin{equation}
\label{eq:redundancy_def}
L_{\text{red}}(y, y_{\text{ref}}) = T_{\text{think}} - t_{\text{anc}}(y, y_{\text{ref}})
\end{equation}
For comparability across responses of varying lengths, we report the redundancy ratio $\rho$:
\begin{equation}
    \rho(y,y_{\text{ref}})=\frac{L_{\text{red}}(y, y_{\text{ref}})}{T_{\text{think}}}
\end{equation}

\paragraph{Observation 1: Reliability of Anchor Localization Methods.}
Fig.~\ref{fig:redundancy_1} illustrates the reliability verification of the two localization methods by comparing the distributions of $\rho(y,\hat{y})$ across four LRMs (averaged over 16 samples per problem).
Quantitative analysis reveals that the average redundancy ratios are comparable between the two methods across all datasets, confirming their effectiveness in locating reasoning anchors at the aggregate level.
However, we observe a divergence on more difficult datasets like AIME24/25, where the model-based localization method tends to position the anchor later.
This discrepancy likely stems from the degradation of instruction-following capabilities in the locator model as the context length increases, highlighting a need for improved robustness in model-based localization for long-context scenarios.

\definecolor{qred}{HTML}{F57C6E}
\definecolor{qgreen}{HTML}{84C3B7}
\definecolor{qblue}{HTML}{71B8ED}
\definecolor{qpink}{HTML}{F2A8DA}

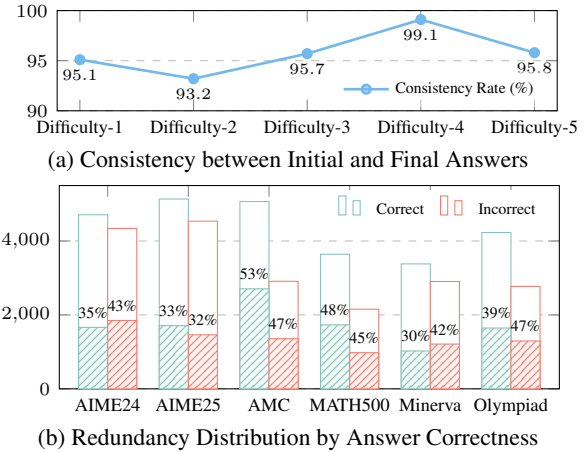
\begin{figure}
    \centering
\hfill      
\begin{tikzpicture}
\begin{axis}[
    width=.48\textwidth,
    height=.17\textwidth,
    ymajorgrids,
    grid style=dashed,
    nodes near coords,
    nodes near coords align={south},
    nodes near coords style={text=black, font=\small, /pgf/number format/fixed, /pgf/number format/fixed zerofill, /pgf/number format/precision=1},
    xmin=0.8, xmax=5.2,
    ymin=90, ymax=100,
    xtick={1,2,3,4,5},
    xticklabels={Difficulty-1,Difficulty-2,Difficulty-3,Difficulty-4,Difficulty-5},
    ytick={90,95,100},
    ticklabel style={font=\scriptsize},
    label style={font=\scriptsize},
    legend style={
        font=\tiny,
        fill opacity=0.5,
        draw=none,
        at={(axis description cs:0.98,0.4)},
        anchor=north east,
        text opacity=1,
    }
]
\addplot[qblue, mark=*,mark size=1.6pt, line width=1pt,nodes near coords,every node near coord/.style={  
    font=\tiny,               
    color=black,               
    anchor=north,
    rotate=0,                  
},
] coordinates {
    (1,95.1) (2,93.2) (3,95.7) (4,99.1) (5,95.8)
};

\legend{Consistency Rate (\%)}
\end{axis}
\end{tikzpicture}

\vspace{-0.4em}
{\small \quad (a) Consistency between Initial and Final Answers}

\vspace{0.4em}

\begin{tikzpicture}
\begin{axis}[
    ybar,
    bar width=11pt,  
    width=.48\textwidth,
    height=.25\textwidth,
    ymajorgrids,
    grid style=dashed,
    symbolic x coords={AIME24,AIME25,AMC,MATH500,Minerva,Olympiad},
    xtick=data,
    tick align=inside,           
    y tick label style={font=\scriptsize},
    x tick label style={rotate=0, anchor=north, font=\scriptsize},
    enlarge x limits=0.12,
    ymin=0, ymax=5500,
    legend style={
        at={(0.98,0.98)},          
        legend columns=2,
        column sep=4pt,
        anchor=north east,         
        font=\tiny,          
        draw=none,                 
        fill=white,                 
        cells={anchor=west},       
        legend image post style={scale=0.9},  
    },
    legend entries={Correct, Incorrect}, 
]

\addplot[fill=white, draw=qgreen, bar shift=-5.5pt] 
coordinates {(AIME24,4710) (AIME25,5135) (AMC,5068) (MATH500,3642) (Minerva,3380) (Olympiad,4229)};

\addplot[fill=white, draw=qred, bar shift=5.5pt] 
coordinates {(AIME24,4340) (AIME25,4536) (AMC,2910) (MATH500,2154) (Minerva,2904) (Olympiad,2772)};

\addplot[
    draw=none,
    bar shift=-5.5pt,
    postaction={fill=qgreen!10,draw=qgreen},  
    postaction={pattern=north east lines, pattern color=qgreen},  
    nodes near coords,  
    point meta=explicit symbolic,  
    every node near coord/.style={  
        font=\tiny,                
        color=black,               
        rotate=0,                  
        xshift=-5.5pt,                
        yshift=0pt,                
        anchor=south,              
    }
] coordinates {(AIME24,1663)[35\%] (AIME25,1710)[33\%] (AMC,2706)[53\%] (MATH500,1730)[48\%] (Minerva,1024)[30\%] (Olympiad,1645)[39\%]};

\addplot[
    draw=none,
    bar shift=5.5pt,
    postaction={fill=qred!10,draw=qred},
    postaction={pattern=north east lines, pattern color=qred},
    nodes near coords,  
    point meta=explicit symbolic,  
    every node near coord/.style={  
        font=\tiny,                
        color=black,               
        rotate=0,                  
        xshift=5.5pt,                
        yshift=0pt,                
        anchor=south,              
    }
] coordinates {(AIME24,1849)[43\%] (AIME25,1461)[32\%] (AMC,1356)[47\%] (MATH500,978)[45\%] (Minerva,1211)[42\%] (Olympiad,1295)[47\%]};

\end{axis}
\end{tikzpicture}

\vspace{-0.4em}
{\small \quad (b) Redundancy Distribution by Answer Correctness}
\caption{
Statistical analysis of reasoning observations.
}
\label{fig:redundancy_2}
\end{figure}
\paragraph{Observation 2: The ``First Answer is Final'' Phenomenon.}
Given that the observed redundancy ratio is nearly 45\%, we conjectured that the model rarely revises a complete answer.
To investigate whether problem difficulty influences self-correction probability, we analyzed the reasoning traces from DS-1.5B on the MATH500 dataset, which features human-annotated difficulty levels.
Specifically, we employed a powerful LLM (Gemini3-Flash) with the prompt detailed in Appendix~\ref{app:prompt} to extract the sentence where the initial answer first appears in the reasoning trace and compared its consistency with the final answer. 
Remarkably, Fig.~\ref{fig:redundancy_2} (a) reveals that the consistency rate is \textbf{95.78\%} regardless of difficulty, even under such a strict extraction-and-matching protocol.
This validates our premise: once the initial answer is formed, it remains largely unchanged; consequently, the subsequent verification-like tokens contribute negligible marginal information gain.

\paragraph{Observation 3: Redundancy Exists Independently of Correctness.}
We further investigated whether redundancy is exclusive to correct reasoning. 
We generated 16 samples for each problem from DS-1.5B across six datasets and categorized them into correct and incorrect groups based on whether the self-generated final answer matches the ground truth.
As shown in Fig.~\ref{fig:redundancy_2} (b), we found non-trivial redundancy ratios in both cases: the correct group exhibits an average redundancy of 39.8\%, while the incorrect group averages 42.5\%.
This evidence strongly suggests that post-anchor generation is not merely a by-product of correct reasoning. 
Instead, redundancy can persist even when the final answer is wrong, reinforcing our argument that redundancy should be defined not by the first occurrence of the \emph{correct} answer, but by the \textit{stability} of the answer.

\paragraph{Observation 4: Verbose Reasoning Structure Impairs Accuracy via Truncation.}
We generated 16 samples per problem from DS-1.5B across six datasets and analyzed the instances where the closing \texttt{</think>} tag was missing within the 8,192-token budget. For these cases, we used the ground truth as the reference answer to identify redundancy.
The statistical results in Table~\ref{tab:pre} reveal that 23.2\% of these truncated responses contain an answer equivalent to the ground truth in the reasoning trace, with an average redundancy ratio of 60.9\%.
However, due to the model's inertial tendency to generating redundant tokens, the maximum token limit was reached before the final answer could be generated, leading to an evaluation failure.
This suggests that an overly verbose reasoning pattern can negatively impact end-task accuracy under practical length constraints.

\begin{table}[t]
    \centering
    \caption{Statistics for samples whose reasoning traces exclude \texttt{</think>}. \textbf{GT-Match} indicates that the ground-truth answer appears in the trace, and \textbf{Redundancy Ratio} is the redundancy proportion among these GT-Match cases. \textit{Note:} Minerva and Olympiad Bench datasets are omitted because they contain no valid matched samples.}
    \resizebox{0.48\textwidth}{!}{%
    \begin{tabular}{l|llll}
\toprule
\textbf{Datasets}                             & \multicolumn{1}{l}{\textbf{AIME24}} & \multicolumn{1}{l}{\textbf{AIME25}} & \multicolumn{1}{l}{\textbf{AMC}} & \multicolumn{1}{l}{\textbf{MATH500}} \\
\midrule
\#No \texttt{</think>} & 351/480                                 & 329/480                                 & 564/1328                              & 525/8000                                  \\
GT-Match Ratio                        & 11.40\%                             & 9.42\%                              & 41.31\%                          & 30.67\%                              \\
Redundancy Ratio                      & 44.90\%                             & 72.20\%                             & 64.50\%                          & 61.80\%                             \\
\bottomrule
\end{tabular}}
    \label{tab:pre}
\end{table}

\paragraph{Conclusion.}
In summary, our analysis demonstrates that redundancy is a common phenomenon regardless of answer correctness or truncation status. 
We reframe overthinking as an intrinsic structural redundancy: while the pre-anchor phase comprises information-dense derivation, correction, and the initial formulation of the conclusion, the post-anchor phase devolves into repetitive verification that rarely revises the answer, thereby contributing negligible marginal information.
However, we do not claim that verification is universally unhelpful. 
We further distinguish two types of verification: \emph{process verification}, which validates intermediate steps during derivation before a complete answer is formed, and \emph{outcome verification}, which redundantly re-checks an already established conclusion. Empirically, the post-anchor phase is dominated by outcome verification. We term this redundant suffix the \emph{Answer-Stable Tail (AST)}.

Motivated by these observations, we seek to design a structure-aware reward that provides precise negative feedback signals targeting this Answer-Stable Tail, thereby incentivizing the model to stop generation upon reaching a confident solution.

\section{Methodology}\label{sec:Methodology}
To achieve this goal, in this section, we first formalize the performance-efficiency trade-off in reasoning models as a multi-objective optimization problem.
Then, we outline the evolution from response-level supervision to our proposed structure-aware \textbf{Anchor-based Process Reward (APR)}. 
Finally, we discuss the selection of policy optimization algorithms, focusing on the theoretical advantages of DAPO over GRPO in handling length penalties.

\subsection{Problem Formulation}\label{4.1}
Our core objective is to train an LRM that mitigates reasoning redundancy to achieve low-latency inference without sacrificing performance.
We formalize this goal as a multi-objective optimization problem, seeking to maximize the joint objective $\mathcal{J}(\theta)$:
\begin{equation}
\label{eq:optimization}
\text{Maximize:} \quad \mathcal{J}(\theta) = \underbrace{\mathcal{P}(\pi_\theta)}_{\text{Performance} \uparrow} - \lambda \cdot \underbrace{\mathcal{C}(\pi_\theta)}_{\text{Latency} \downarrow}
\end{equation}
where $\mathcal{P}(\cdot)$ represents the model's performance (e.g., accuracy) and $\mathcal{C}(\cdot)$ denotes the inference efficiency (e.g., token count). The coefficient $\lambda \ge 0$ serves as a hyperparameter controlling the model's preference between sufficient reasoning and concise generation.

\subsection{Anchor-based Process Reward}\label{4.2}
We adopt the RLVR paradigm to instantiate the aforementioned objective defined in Eq.~\ref{eq:optimization}. 

\paragraph{Reward with Length Penalty.}
To incorporate the efficiency constraint (i.e., $\mathcal{C}(\pi_\theta)$), a conventional approach augments the accuracy reward with a penalty proportional to the generated sequence length:
\begin{equation}
\label{eq:length}
R_{\text{len}}(y) = \mathbb{I}(\hat{y} = y^*) - \beta \cdot L(y)
\end{equation}
where $\mathbb{I}(\cdot)$ is the correctness indicator function, $\beta$ is the penalty coefficient and $L(y)$ denotes the length reward, which varies across different methods.
However, this coarse-grained penalty applies uniformly to the entire sequence, making no distinction between necessary derivation and redundant verification.
As a result, it provides a misaligned training signal: the model is indiscriminately discouraged from generating tokens, rather than being specifically guided to suppress the truly redundant segment.

To achieve a more granular trade-off between reasoning performance and efficiency, we propose to replace the response-level length penalty with a structure-aware process supervision signal.
Motivated by the informational contrast between reasoning phases, we design \textbf{Anchor-based Process Reward (APR)} to penalize the redundant Answer-Stable Tail (AST) while preserving the necessary pre-anchor phases.
We first identify the reasoning anchor $t_{\text{anc}}(y,\hat{y})$ using Eq.~\ref{eq:anchor_func} and use $L_{\text{AST}}(y,y_{\text{ref}})$ to refer to the redundancy length defined in Eq.~\ref{eq:redundancy_def}.
Following the setting in Sec.~\ref{sec:empirical_analysis}, we focus on \emph{complete} responses that contain the closing tag \texttt{</think>} and use the self-generated final answer $\hat{y}$ as the reference to localize redundancy.

\paragraph{Anchor-based Process Reward (APR).}
To preserve end-task correctness, we penalize the AST length only when the final answer is correct:
\begin{equation}
\label{eq:apr_acc_first}
R_{\text{APR}}(y)
= \mathbb{I}(\hat{y}=y^*)\cdot\big(1-\beta\,L_{\text{AST}}(y,\hat{y})\big)
\end{equation}
where $\mathbb{I}(\cdot)$ is the indicator function and $\beta$ controls the strength of the redundancy penalty.
This zero lower bound prevents a misalignment where incorrect complete responses receive negative rewards lower than truncated rollouts, which would incentivize the model to prefer truncation.

\subsection{Policy Optimization: From GRPO to DAPO}\label{4.3}
While $R_{\text{APR}}$ can be integrated into a range of policy optimization paradigms, the optimization algorithm can meaningfully affect how length-related signals are translated into stable learning updates.
In this section, we discuss a normalization-related limitation that may arise when applying standard Group Relative Policy Optimization (GRPO)~\cite{shao2024deepseekmath} to length-aware rewards, and then introduce Direct Alignment Policy Optimization (DAPO)~\cite{yu2025dapo} as a practical alternative.

\paragraph{A limitation of GRPO under length-aware rewards.}
GRPO estimates the group-normalized advantage by standardizing rewards within a group:
$\hat{A}_i = \frac{R_i - \mu_R}{\sigma_R + \epsilon}$,
where $\mu_R$ and $\sigma_R$ are the mean and standard deviation of the rewards $\{R_i\}_{i=1}^{G}$ within the group, and $\epsilon>0$ is a small stabilizer.
While such normalization is widely used and generally effective for binary outcome signals, it can reduce the effective sensitivity to the penalty coefficient $\beta$ in homogeneous groups.
Consider a \textit{Correct-but-Verbose} scenario where a group consists entirely of correct responses ($\hat{y}=y^*$, i.e., $\mathbb{I}=1$) that differ only in their AST length.

Recall $R_{\text{APR}}(y)=1-\beta L_{\text{AST}}(y,\hat{y})$ for these correct responses.
In a fully correct group, the constant term is removed by mean subtraction, yielding:
\begin{equation}
\label{eq:grpo_fail_additive_eps}
\hat{A}_i
= \frac{-\beta(L_i-\mu_L)}{\beta\sigma_L+\epsilon}
\end{equation}
where $L_i$ denotes $L_{\text{AST}}$ for the $i$-th rollout and $\mu_L,\sigma_L$ are the mean and standard deviation of $\{L_i\}$ within the group.
When the within-group variation is dominated by the length term (i.e., $\beta\sigma_L \gg \epsilon$), Eq.~\ref{eq:grpo_fail_additive_eps} reduces to:
\begin{equation}
\label{eq:grpo_fail_additive_approx}
\hat{A}_i \approx -\frac{L_i-\mu_L}{\sigma_L}
\end{equation}
This derivation indicates that, in this regime, the normalized optimization signal becomes approximately independent of $\beta$.
As a result, tuning $\beta$ may have limited effect on the strength of the length preference within such homogeneous groups, which can make the accuracy-efficiency trade-off harder to calibrate in practice.

\paragraph{DAPO as an alternative.}
To mitigate this normalization issue in homogeneous batches, we employ DAPO as an alternative optimization choice.
DAPO introduces a dynamic sampling mechanism (Appendix~\ref{app:algo}) that filters for groups with correctness and performs policy updates using only groups that contain both correct and incorrect responses:
\begin{equation}
\label{eq:dapo_constraint}
\begin{split}
    \mathcal{J}_{\mathrm{DAPO}}(\theta) &= \mathbb{E}_{(q,a)\sim \mathcal{D}, \{o_i\}_{i=1}^G\sim \pi_{\theta_\text{old}}(\cdot\mid q)} \left[ \dots \right] \\
    &\text{s.t. } 0 < |\{o_i \mid \text{is\_correct}(o_i)\}| < G
\end{split}
\end{equation}
By excluding homogeneous groups, specifically the \textit{Correct-but-Verbose} scenarios where reward variance is purely length-dependent, DAPO bypasses the degenerate regime where the penalty coefficient $\beta$ is neutralized by normalization.
Under this formulation, the length penalty functions as a consistent margin modulator within a stable optimization landscape dominated by correctness signals, rather than acting as the sole, volatile source of variance.
This ensures that $\beta$ remains a controllable hyperparameter for tuning the accuracy-efficiency trade-off.

\begin{table*}[t]
    \centering
    \caption{Comparison of performance and efficiency across five common mathematical reasoning benchmarks.We generate 16 response samples per problem to report the average Pass@1 accuracy (\%), average generation length (tokens), and AE Score. The best results are highlighted in \textbf{bold} and the second-best in \underline{underlined}.}
    \resizebox{0.99\textwidth}{!}{%
    {
\setlength{\tabcolsep}{3pt}

\begin{tabular}{lcccccccccccccccccc}

\toprule
\multicolumn{1}{c}{\multirow{2.5}{*}{\textbf{Model}}} & 
\multicolumn{3}{c}{\textbf{AIME24}} & 
\multicolumn{3}{c}{\textbf{AMC}} & 
\multicolumn{3}{c}{\textbf{MATH500}} & 
\multicolumn{3}{c}{\textbf{Minerva}} & 
\multicolumn{3}{c}{\textbf{Olympiad\_bench}} & 
\multicolumn{3}{c}{\textbf{Overall Performance}} \\

\cmidrule(lr){2-4} \cmidrule(lr){5-7} \cmidrule(lr){8-10} \cmidrule(lr){11-13} \cmidrule(lr){14-16} \cmidrule(lr){17-19}

\multicolumn{1}{c}{} & 
\makecell{Avg.\\@16} & \makecell{Avg.\\Tokens} & \makecell{AE $\uparrow$\\Score} & 
\makecell{Avg.\\@16} & \makecell{Avg.\\Tokens} & \makecell{AE $\uparrow$\\Score} & 
\makecell{Avg.\\@16} & \makecell{Avg.\\Tokens} & \makecell{AE $\uparrow$\\Score} & 
\makecell{Avg.\\@16} & \makecell{Avg.\\Tokens} & \makecell{AE $\uparrow$\\Score} & 
\makecell{Avg.\\@16} & \makecell{Avg.\\Tokens} & \makecell{AE $\uparrow$\\Score} & 
\makecell{$\Delta$ (\%)\\Acc.} & \makecell{$\Delta$ (\%)\\Tokens} & \makecell{AE $\uparrow$\\Score} \\ 
\midrule

\rowcolor{gray!10} \multicolumn{19}{c}{\textit{\textbf{Based on DeepSeek-R1-Distill-Qwen-1.5B}}} \\ 
\midrule
Original Model & 19.2 & 7278 & \multicolumn{1}{c}{-} & 51.5 & 5727 & \multicolumn{1}{c}{-} & \underline{85.1} & 3112 & \multicolumn{1}{c}{-} & 30.3 & 4082 & \multicolumn{1}{c}{-} & 37.5 & 5775 & \multicolumn{1}{c}{-} & \multicolumn{1}{c}{-} & \multicolumn{1}{c}{-} & \multicolumn{1}{c}{-} \\
AdaptThink-1.5B-delta0.05 & 26.2 & 6172 & 1.25 & 61.7 & 3428 & 1.00 & 80.8 & \underline{1532} & 0.26 & 24.7 & \textbf{1637} & -0.33 & 40.1 & 3612 & 0.58 & 4.4\% & 36.9\% & 0.50 \\
L1-Qwen-1.5B-Max & 28.8 & \textbf{2893} & \underline{2.10} & 67.8 & \textbf{2300} & 1.55 & 84.8 & 1899 & 0.37 & 29.4 & 2779 & 0.17 & 46.2 & \textbf{2311} & \underline{1.30} & 14.9\% & \textbf{53.1\%} & 0.98 \\
TrainingEfficient\_DS-1.5B & 29.2 & 6245 & 1.70 & 58.9 & 4231 & 0.69 & 81.8 & 2285 & 0.07 & 27.0 & 3018 & -0.28 & 40.9 & 4523 & 0.49 & 6.4\% & 21.8\% & 0.41 \\
DS-1.5B-thinkprune-iter2k & \underline{30.0} & 4670 & 2.05 & 65.2 & 2964 & 1.28 & 83.1 & 1822 & 0.30 & 27.6 & 2118 & 0.04 & 42.9 & 3162 & 0.88 & 11.3\% & 43.3\% & 0.77 \\
Laser-L8192-1.5B & 27.3 & 6139 & 1.42 & 68.3 & 4212 & 1.24 & 84.9 & 2608 & 0.15 & \underline{31.1} & 3638 & 0.19 & \underline{47.1} & 4076 & 1.06 & 15.7\% & 20.4\% & 0.68 \\
Laser-DE-L4096-1.5B & 25.6 & 5173 & 1.29 & 65.9 & 3333 & 1.26 & 82.8 & 1909 & 0.25 & 29.0 & 2289 & 0.22 & 44.2 & 3335 & 0.96 & 10.7\% & 38.2\% & 0.70 \\
DLER-R1-1.5B-Research & \textbf{32.1} & \underline{3376} & \textbf{2.55} & \textbf{74.2} & 2559 & \textbf{1.88} & \textbf{86.9} & 1787 & \textbf{0.49} & \textbf{31.5} & 2225 & \textbf{0.57} & \textbf{49.7} & 2595 & \textbf{1.53} & \textbf{22.7\%} & 51.7\% & \textbf{1.20} \\
\textbf{APR-1.5B (Ours)} & 29.0 & 3767 & 2.01 & \underline{69.6} & \underline{2532} & \underline{1.61} & 84.7 & \textbf{1513} & \textbf{0.49} & 30.2 & \underline{1985} & \underline{0.50} & 46.4 & \underline{2450} & 1.29 & \underline{16.3\%} & \underline{52.8\%} & \underline{1.02} \\
\midrule

\rowcolor{gray!10} \multicolumn{19}{c}{\textit{\textbf{Based on DeepSeek-R1-Distill-Qwen-7B}}} \\ 
\midrule
Original Model & 37.7 & 6707 & \multicolumn{1}{c}{-} & 69.5 & 5014 & \multicolumn{1}{c}{-} & 86.7 & 3274 & \multicolumn{1}{c}{-} & 36.2 & 4266 & \multicolumn{1}{c}{-} & 46.1 & 5395 & \multicolumn{1}{c}{-} & \multicolumn{1}{c}{-} & \multicolumn{1}{c}{-} & \multicolumn{1}{c}{-} \\
AdaptThink-7B-delta0.05 & 45.6 & 6255 & 0.70 & 75.2 & 4103 & 0.43 & 88.2 & 1946 & 0.46 & 35.1 & 2570 & 0.25 & 50.8 & 4402 & 0.49 & 6.7\% & 21.8\% & 0.42 \\
L1-Qwen-7B-Max & 44.8 & 4041 & 0.96 & 77.4 & 2742 & 0.79 & 90.2 & 2125 & 0.47 & \underline{38.9} & 2120 & 0.73 & 52.5 & 2835 & 0.89 & 10.0\% & 43.8\% & 0.74 \\
TrainingEfficient\_DS-7B & 42.1 & 6230 & 0.42 & 75.3 & 4169 & 0.42 & 89.1 & 2427 & 0.34 & 37.7 & 2940 & 0.44 & 51.8 & 4437 & 0.55 & 7.2\% & 18.1\% & 0.40 \\
SB\_DS7B\_alpha\_2 & 45.4 & 3955 & 1.02 & 74.4 & 2272 & 0.76 & 82.6 & \textbf{1037} & 0.45 & 31.6 & \textbf{901} & 0.15 & 49.6 & 2361 & 0.79 & 2.7\% & \textbf{57.3\%} & 0.65 \\
Laser-DE-L4096-7B & \underline{47.7} & 4642 & \underline{1.10} & \underline{82.1} & 2793 & 0.99 & \underline{91.5} & 1634 & \underline{0.67} & \underline{38.9} & 1850 & \underline{0.79} & \underline{56.0} & 2998 & 1.09 & \underline{14.5\%} & 43.6\% & 0.87 \\
DLER-R1-7B-Research & \textbf{51.3} & \underline{3209} & \textbf{1.60} & \textbf{83.3} & \textbf{2230} & \textbf{1.15} & \textbf{91.8} & \underline{1429} & \textbf{0.74} & \textbf{39.5} & 1798 & \textbf{0.85} & \textbf{57.2} & \underline{2316} & \textbf{1.29} & \textbf{17.0\%} & 55.5\% & \textbf{1.06} \\
\textbf{APR-7B (Ours)} & 44.0 & \textbf{3051} & 1.04 & 81.4 & \underline{2256} & \underline{1.06} & 90.3 & 1494 & \underline{0.67} & 38.4 & \underline{1647} & \underline{0.79} & 54.0 & \textbf{2235} & \underline{1.10} & 11.5\% & \underline{56.7\%} & \underline{0.91} \\
\bottomrule
\end{tabular}

}}
    \label{tab:main}
\end{table*}

\section{Experiment}

\subsection{Setup}
\paragraph{Implementation Details.}
We select DeepSeek-R1-Distill-Qwen-1.5B and 7B~\cite{guo2025deepseek} as our backbone LRMs, given their widespread adoption as strong baselines.
The training dataset is the DeepScaleR-preview~\cite{deepscaler2025}, which comprises 40K mathematical problems drawn from the AIME, AMC, Omni-MATH, and Still datasets. Guided by our preliminary findings in Section~\ref{3.2}, we removed samples where rule-based methods struggle to identify the correct anchor, resulted in a curated subset of 33,113 samples.
We implement our method based on \textbf{VeRL}~\cite{sheng2024hybridflow}, an open-source RL training library for post-training.
We adhere to the original prompt template from DeepSeek-R1 to ensure fair comparison (see Appendix~\ref{app:prompt}).
The training configuration includes a global batch size of $N=128$, a rollout group size of $n=8$, and a maximum sequence length of 8,192 tokens. A comprehensive list of hyperparameters is provided in Appendix~\ref{app:hyperparameters}.

\paragraph{Baselines.}
To validate the effectiveness of APR, we compare our method against six widely-used, open-source efficient-reasoning models detailed in Appendix~\ref{app:baselines}.
All results are derived from evaluating the officially released checkpoints under a unified experimental setup.

\paragraph{Evaluation.}
We evaluate the average Pass@1 accuracy and generation length over 16 samples, as well as the AE score, an inference efficiency metric detailed in Appendix~\ref{app:ae_score}, across five datasets: AIME24~\cite{AIME}, AMC~\cite{AMC}, MATH500~\cite{hendrycks2021measuring}, Minerva~\cite{lewkowycz2022solving}, and Olympiad Bench~\cite{he2024olympiadbench}. AIME25 results are omitted due to their high consistency with AIME24 performance.
All evaluations are conducted using vLLM~\cite{kwon2023efficient} as the inference backend with a sampling temperature of $0.6$ and a maximum response length of 8,192 tokens.

\subsection{Main Results}
\paragraph{Pareto Frontier of Performance-Efficiency.} 
Visualizing the accuracy-length trade-offs averaged across five datasets in Fig.~\ref{fig:pareto} in Appendix~\ref{app:addition} reveals that our APR model achieves the Pareto frontier across both 1.5B and 7B scales.
Table~\ref{tab:main} provides a detailed report on the accuracy, generation length, and AE scores for all compared baselines across the datasets.
In terms of inference efficiency, our APR model consistently ranks first or second in generation length across all evaluated benchmarks, except for AIME24.
In comparison, L1-Qwen-1.5B-Max, which rivals our model in generation length, underperforms in terms of accuracy.
Such findings demonstrate that the APR method effectively reduces the generation length of LRMs, thereby enhancing inference efficiency. 
Furthermore, APR maintains competitive reasoning performance relative to other models. 
While it may not strictly dominate every individual dataset, APR ranks second in terms of average performance across the five benchmarks, suggesting strong robustness across diverse tasks.
We also calculated the AE scores based on average accuracy and length, where our model shows clear advantages on all datasets excluding AIME24, highlighting an excellent balance between performance and efficiency.

\begin{table}[t]
    \centering
    \caption{Average thinking length and redundancy ratio for 1.5B models across five datasets. \textbf{Bold}: best; \underline{underlined}: second-best. See detailed Table~\ref{tab:redundancy_ratio} in Appendix~\ref{app:addition}.}
    \resizebox{0.48\textwidth}{!}{%
    {

\begin{tabular}{llllllll} 
\toprule
\textbf{1.5B Models} & 
\makecell{L1\\-Max} & 
\makecell{Training\\Efficient} & 
\makecell{Think\\Prune} & 
\makecell{Laser\\-DE} & 
\makecell{DLER\\-R1} & 
\makecell{APR\\-1.5B} \\ 
\midrule
Thinking Length $\downarrow$ & 
\underline{2102} & 2320 & 
2146 & 2417 & 2129 & \textbf{2011} \\
Redundancy Ratio $\downarrow$ & 
27.8\% & 21.9\% & 
\underline{20.2\%} & 20.3\% & 23.0\% & \textbf{14.2\%} \\ 
\bottomrule
\end{tabular}

}}
    \label{tab:ratio}
\end{table}

\paragraph{Substantial Reduction of Structural Redundancy.}
To investigate the source of length reduction, we leveraged the rule-based anchor localization method proposed in Section~\ref{sec:preliminary} to quantify the Answer-Stable Tail using 16 samples from our best-performing 1.5B and 7B models, as shown in Table~\ref{tab:ratio}. 
Compared to all baselines, our model exhibits the lowest length of thinking process and AST redundancy ratio.
These results confirm that the APR method specifically targets and reduces the proportion of AST redundancy, with the anchor-based process reward playing a pivotal role in efficient reasoning.
The comparison further implies that global length penalties lack the granularity to effectively eliminate structural redundancy, as they fail to provide precise penalization for the truly redundant segments.

\paragraph{Dense Rewards Facilitate Efficient Training.}
Computational resources during RL training are a critical bottleneck for LRM deployment.
Taking the 1.5B model as a case study, we compared the training resource consumption with L1-Qwen-1.5-Max, selected for its similar AE score and the availability of detailed training specifications.
It is important to note that their starting point, the base LRM (DeepScaleR-1.5B-Preview), is already a strong checkpoint that has undergone 1,750 steps (3,800 A100 hours) of training on top of DS-1.5B (our base LRM).
Under identical settings except for training steps and rollout counts (see Appendix~\ref{app:hyperparameters}), L1 requires a total of 820 steps (700 for stage 1 and 120 for stage 2), while our best model is achieved after only 250 steps of DAPO training with half the rollout count of L1.
Despite starting from a weaker base LRM and consuming fewer resources, APR achieves comparable or superior overall performance.
We attribute this high training efficiency to the dense and accurate feedback signals provided by the process reward, which guide policy optimization more effectively than sparse outcome rewards.

\subsection{Ablation Study}
\paragraph{Comparison of Anchor Localization Methods.} 
To investigate the divergence between Rule-based and Model-based localization methods, we employed two parallel experimental settings using the DAPO: an 1.5B model with $\beta = 2\text{e-}4$ trained for 250 steps and a 7B model with $\beta = 5\text{e-}4$ trained for 100 steps.
As shown in Fig.~\ref{fig:localization} in Appendix~\ref{app:localization}, our results indicate that the rule-based method consistently outperforms the model-based approach in terms of accuracy with a slight advantage, while yielding comparable results in generation length.
Notably, the performance gap becomes more pronounced on the AIME24 dataset which corroborates Observation 1 in Section~\ref{sec:empirical_analysis}, suggesting that the model-based locator is prone to ``lost-in-the-middle'' hallucinations when identifying anchors within long contexts. 
As a result, our optimal model utilizes the rule-based anchor localization method to determine the length of AST.

\paragraph{Sensitivity Analysis of Penalty Coefficient $\beta$.}
To determine $\beta$, we analyzed the redundancy distribution of sampled training trajectories, which revealed an average redundant length ($L_\text{avg}$) of $\approx 1,400$ tokens.
To ensure positive initial rewards ($1 - \beta \cdot L_\text{avg} > 0$), we derived an upper bound of $\beta < 6\text{e-}4$ and selected $\beta \in \{2\text{e-}4, 5\text{e-}4\}$.
We compared two penalty coefficients using the DS-1.5B model trained with GRPO for 250 steps, observing obvious performance divergences.
Specifically, the $\beta = 2\text{e-}4$ setting improves accuracy by 4.0\% higher than the $5\text{e-}4$ setting, but with a 4.4\% smaller reduction in generation length.
This indicates that the length penalty coefficient serves as a critical factor for balancing the trade-off between performance and efficiency in LRMs.
Our reported 1.5B model is trained with $\beta = 2\text{e-}4$, and the 7B model with $\beta = 5\text{e-}4$.

\paragraph{Choice of Policy Optimization: DAPO vs. GRPO.} 
Based on the DS-1.5B model, we compared the two algorithms in terms of average accuracy and generation length.
For accuracy improvement ($\beta = 2\text{e-}4$), GRPO requires 250 steps to match the performance of DAPO at 50 steps.
For length reduction ($\beta = 5\text{e-}4$), the disparity is more pronounced: GRPO needs 600 steps to achieve the compactness DAPO reaches in just 50 steps.
We attribute DAPO's efficiency to its soft overlong punishment, which effectively mitigates exploration issues during the early training stages.
However, the effectiveness of the APR penalty itself is validated by GRPO's subsequent behavior.
Between steps 50 and 100, GRPO exhibits a rapid 13.6\% length reduction, a rate that actually surpasses that of DAPO.
This confirms that the APR penalty provides a consistently strong gradient signal for redundancy elimination, while DAPO serves to unlock this potential more efficiently in the initial phase.

\section{Conclusion}
In this work, we conceptually rethink the essence of overthinking in LRMs as originating from intrinsic structural redundancy.
We introduce the Reasoning Anchor to localize the repetitive verification phase following answer stabilization, termed the Answer-Stable Tail (AST).
Building on this, we propose the Anchor-based Process Reward method combined with the DAPO policy optimization algorithm, achieving the Pareto frontier across five mathematical reasoning datasets while utilizing fewer training resources.



\section*{Acknowledgements}
This work was supported in part by the National Science Foundation of China (Nos. 62276056 and U24A20334), the Yunnan Fundamental Research Projects (No.202401BC070021), the Yunnan Science and Technology Major Project (No. 202502AD080014), the Fundamental Research Funds for the Central Universities (Nos. N25BSS054 and N25BSS094), and the Program of Introducing Talents of Discipline to Universities, Plan 111 (No.B16009).

\section*{Impact Statement}
This paper presents work whose goal is to advance the field of Machine
Learning. There are many potential societal consequences of our work, none
which we feel must be specifically highlighted here.


\bibliography{example_paper}
\bibliographystyle{icml2026}

\newpage
\appendix
\onecolumn

\section{Additional Experimental Results}
\label{app:addition}
\subsection{Pareto Frontier}
We present the performance-efficiency Pareto frontier averaged across five datasets in Fig.~\ref{fig:pareto}. Both APR-1.5B and APR-7B successfully achieve the Pareto frontier. Notably, APR-1.5B shortens generation length by up to 52.8\% while improving accuracy by 16.3\%, and APR-7B achieves a 56.7\% length reduction accompanied by a 11.5\% accuracy gain.

\definecolor{qred}{HTML}{F57C6E}
\definecolor{qgreen}{HTML}{84C3B7}
\definecolor{qblue}{HTML}{71B8ED}
\definecolor{qpink}{HTML}{F2A8DA}
\definecolor{qpurple}{HTML}{B39DDB}
\definecolor{qorange}{HTML}{FFB74D}
\definecolor{qteal}{HTML}{50749E}
\definecolor{qbrown}{HTML}{A1887F}
\definecolor{qgray}{HTML}{90A4AE}

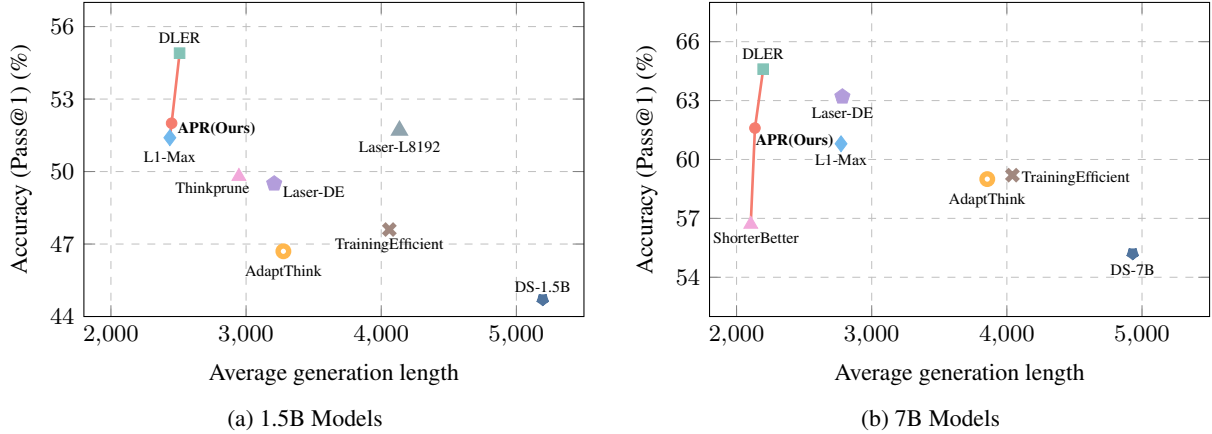
\begin{figure}[h]
    \centering
    
\begin{minipage}[t]{0.48\textwidth}
\centering    
\begin{tikzpicture}
\begin{axis}[
    width=\textwidth,
    height=0.7\textwidth,
    ymajorgrids,
    xmajorgrids=true,
    grid style={dashed, gray!50, line width=0.3pt},
    xmin=1800, xmax=5500,
    ymin=44, ymax=57,
    xtick={2000,3000,4000,5000},
    ytick={44,47,50,53,56},
    ticklabel style={font=\footnotesize},
    label style={font=\footnotesize},
    xlabel={Average generation length},
    ylabel={Accuracy (Pass@1) (\%)},
]


\addplot[
    qred, 
    mark=*,
    mark size=2pt,
    line width=0,
    only marks,
    nodes near coords,
    point meta=explicit symbolic,
    visualization depends on={value \thisrow{label} \as \mylabel},
    every node near coord/.style={
        font=\tiny,
        color=black,
        anchor=center,
        xshift=17pt,
        yshift=-2pt,
    }
] table [meta=label] {
    x y label
    2449 52 \textbf{APR(Ours)}
};

\addplot[
    qgreen, 
    mark=square*,
    mark size=2pt,
    line width=0,
    only marks,
    nodes near coords,
    point meta=explicit symbolic,
    visualization depends on={value \thisrow{label} \as \mylabel},
    every node near coord/.style={
        font=\tiny,
        color=black,
        anchor=center,
        xshift=0pt,
        yshift=6pt,
    }
] table [meta=label] {
    x y label
    2508 54.9 DLER
};

\addplot[
    qblue, 
    mark=diamond*,
    mark size=3pt,
    line width=0,
    only marks,
    nodes near coords,
    point meta=explicit symbolic,
    visualization depends on={value \thisrow{label} \as \mylabel},
    every node near coord/.style={
        font=\tiny,
        color=black,
        anchor=center,
        xshift=0pt,
        yshift=-7pt,
    }
] table [meta=label] {
    x y label
    2436 51.4 L1-Max
};

\addplot[
    qpink, 
    mark=triangle*,
    mark size=3pt,
    line width=0,
    only marks,
    nodes near coords,
    point meta=explicit symbolic,
    visualization depends on={value \thisrow{label} \as \mylabel},
    every node near coord/.style={
        font=\tiny,
        color=black,
        anchor=center,
        xshift=-10pt,
        yshift=-5pt,
    }
] table [meta=label] {
    x y label
    2947 49.8 Thinkprune
};

\addplot[
    qpurple, 
    mark=pentagon*,
    mark size=3pt,
    line width=0,
    only marks,
    nodes near coords,
    point meta=explicit symbolic,
    visualization depends on={value \thisrow{label} \as \mylabel},
    every node near coord/.style={
        font=\tiny,
        color=black,
        anchor=center,
        xshift=15pt,
        yshift=-3pt,
    }
] table [meta=label] {
    x y label
    3208 49.5 Laser-DE
};

\addplot[
    qorange, 
    mark=o,
    mark size=2pt,
    line width=2pt,
    only marks,
    nodes near coords,
    point meta=explicit symbolic,
    visualization depends on={value \thisrow{label} \as \mylabel},
    every node near coord/.style={
        font=\tiny,
        color=black,
        anchor=center,
        xshift=0pt,
        yshift=-8pt,
    }
] table [meta=label] {
    x y label
    3276 46.7 AdaptThink
};

\addplot[
    qteal, 
    mark=star,
    mark size=2pt,
    line width=3pt,
    only marks,
    nodes near coords,
    point meta=explicit symbolic,
    visualization depends on={value \thisrow{label} \as \mylabel},
    every node near coord/.style={
        font=\tiny,
        color=black,
        anchor=center,
        xshift=0pt,
        yshift=5pt,
    }
] table [meta=label] {
    x y label
    5195 44.7 DS-1.5B
};

\addplot[
    qbrown, 
    mark=x,
    mark size=3pt,
    line width=2pt,
    only marks,
    nodes near coords,
    point meta=explicit symbolic,
    visualization depends on={value \thisrow{label} \as \mylabel},
    every node near coord/.style={
        font=\tiny,
        color=black,
        anchor=center,
        xshift=0pt,
        yshift=-6pt,
    }
] table [meta=label] {
    x y label
    4060 47.6 TrainingEfficient
};

\addplot[
    qgray, 
    mark=triangle,
    mark size=2pt,
    line width=2pt,
    only marks,
    nodes near coords,
    point meta=explicit symbolic,
    visualization depends on={value \thisrow{label} \as \mylabel},
    every node near coord/.style={
        font=\tiny,
        color=black,
        anchor=center,
        xshift=0pt,
        yshift=-6pt,
    }
] table [meta=label] {
    x y label
    4135 51.7 Laser-L8192
};

\addplot[
    draw=qred,          
    line width=1pt,     
    no markers          
] coordinates {
    (2436, 51.4)
    (2449, 52)
    (2508, 54.9)
};

\end{axis}
\end{tikzpicture}

{\small \quad (a) 1.5B Models}
\end{minipage}
\begin{minipage}[t]{0.48\textwidth}
\centering
\begin{tikzpicture}
\begin{axis}[
    width=\textwidth,
    height=0.7\textwidth,
    ymajorgrids,
    xmajorgrids=true,
    grid style={dashed, gray!50, line width=0.3pt},
    xmin=1800, xmax=5500,
    ymin=52, ymax=68,
    xtick={2000,3000,4000,5000},
    ytick={54,57,60,63,66},
    ticklabel style={font=\footnotesize},
    label style={font=\footnotesize},
    xlabel={Average generation length},
    ylabel={Accuracy (Pass@1) (\%)},
]


\addplot[
    qred, 
    mark=*,
    mark size=2pt,
    line width=0,
    only marks,
    nodes near coords,
    point meta=explicit symbolic,
    visualization depends on={value \thisrow{label} \as \mylabel},
    every node near coord/.style={
        font=\tiny,
        color=black,
        anchor=center,
        xshift=15pt,
        yshift=-5pt,
    }
] table [meta=label] {
    x y label
    2136 61.6 \textbf{APR(Ours)}
};

\addplot[
    qgreen, 
    mark=square*,
    mark size=2pt,
    line width=0,
    only marks,
    nodes near coords,
    point meta=explicit symbolic,
    visualization depends on={value \thisrow{label} \as \mylabel},
    every node near coord/.style={
        font=\tiny,
        color=black,
        anchor=center,
        xshift=0pt,
        yshift=6pt,
    }
] table [meta=label] {
    x y label
    2197 64.6 DLER
};

\addplot[
    qblue, 
    mark=diamond*,
    mark size=3pt,
    line width=0,
    only marks,
    nodes near coords,
    point meta=explicit symbolic,
    visualization depends on={value \thisrow{label} \as \mylabel},
    every node near coord/.style={
        font=\tiny,
        color=black,
        anchor=center,
        xshift=0pt,
        yshift=-6pt,
    }
] table [meta=label] {
    x y label
    2773 60.8 L1-Max
};

\addplot[
    qpink, 
    mark=triangle*,
    mark size=3pt,
    line width=0,
    only marks,
    nodes near coords,
    point meta=explicit symbolic,
    visualization depends on={value \thisrow{label} \as \mylabel},
    every node near coord/.style={
        font=\tiny,
        color=black,
        anchor=center,
        xshift=2pt,
        yshift=-5pt,
    }
] table [meta=label] {
    x y label
    2105 56.7 ShorterBetter
};

\addplot[
    qpurple, 
    mark=pentagon*,
    mark size=3pt,
    line width=0,
    only marks,
    nodes near coords,
    point meta=explicit symbolic,
    visualization depends on={value \thisrow{label} \as \mylabel},
    every node near coord/.style={
        font=\tiny,
        color=black,
        anchor=center,
        xshift=0pt,
        yshift=-6pt,
    }
] table [meta=label] {
    x y label
    2783 63.2 Laser-DE
    
};

\addplot[
    qorange, 
    mark=o,
    mark size=2pt,
    line width=2pt,
    only marks,
    nodes near coords,
    point meta=explicit symbolic,
    visualization depends on={value \thisrow{label} \as \mylabel},
    every node near coord/.style={
        font=\tiny,
        color=black,
        anchor=center,
        xshift=0pt,
        yshift=-8pt,
    }
] table [meta=label] {
    x y label
    3855 59 AdaptThink
};

\addplot[
    qteal, 
    mark=star,
    mark size=2pt,
    line width=3pt,
    only marks,
    nodes near coords,
    point meta=explicit symbolic,
    visualization depends on={value \thisrow{label} \as \mylabel},
    every node near coord/.style={
        font=\tiny,
        color=black,
        anchor=center,
        xshift=0pt,
        yshift=-6pt,
    }
] table [meta=label] {
    x y label
    4931 55.2 DS-7B
};

\addplot[
    qbrown, 
    mark=x,
    mark size=3pt,
    line width=2pt,
    only marks,
    nodes near coords,
    point meta=explicit symbolic,
    visualization depends on={value \thisrow{label} \as \mylabel},
    every node near coord/.style={
        font=\tiny,
        color=black,
        anchor=center,
        xshift=24pt,
        yshift=-1pt,
    }
] table [meta=label] {
    x y label
    4041 59.2 TrainingEfficient
};

\addplot[
    draw=qred,          
    line width=1pt,     
    no markers          
] coordinates {
    (2105, 56.7)
    (2136, 61.6)
    (2197, 64.6)
};

\end{axis}
\end{tikzpicture}

{\small \quad (b) 7B Models}
\end{minipage}

\caption{Performance-efficiency Pareto frontier averaged across five datasets.}
\label{fig:pareto}
\end{figure}

\subsection{Structural Redundancy Ratio}
Table~\ref{tab:redundancy_ratio} presents a detailed comparison of average generation length and AST redundancy ratio for all 1.5B and 7B models across five datasets.

\begin{table*}[h]
    \centering
    \caption{Comparison of structural redundancy across five common mathematical reasoning benchmarks. We generate 16 response samples per problem to report the average length of the thinking process and AST redundancy ratio. The best results are highlighted in \textbf{bold} and the second-best in \underline{underlined}.}
    \resizebox{0.99\textwidth}{!}{%
    \begin{tabular}{lrrrrrrrrrr}
\toprule
\multicolumn{1}{c}{\multirow{2.5}{*}{\textbf{Models}}} & 
\multicolumn{2}{c}{\textbf{AIME24}} & 
\multicolumn{2}{c}{\textbf{AMC}} & 
\multicolumn{2}{c}{\textbf{MATH500}} & 
\multicolumn{2}{c}{\textbf{Minerva}} & 
\multicolumn{2}{c}{\textbf{Olympiad\_Bench}} \\

\cmidrule(lr){2-3} \cmidrule(lr){4-5} \cmidrule(lr){6-7} \cmidrule(lr){8-9} \cmidrule(lr){10-11}

\multicolumn{1}{c}{} & 
\makecell{Thinking\\Length $\downarrow$} & \makecell{Redundancy\\Ratio $\downarrow$} & 
\makecell{Thinking\\Length $\downarrow$} & \makecell{Redundancy\\Ratio $\downarrow$} & 
\makecell{Thinking\\Length $\downarrow$} & \makecell{Redundancy\\Ratio $\downarrow$} & 
\makecell{Thinking\\Length $\downarrow$} & \makecell{Redundancy\\Ratio $\downarrow$} & 
\makecell{Thinking\\Length $\downarrow$} & \makecell{Redundancy\\Ratio $\downarrow$} \\ 
\midrule

\rowcolor{gray!10} \multicolumn{11}{c}{\textit{Based on DeepSeek-R1-Distill-Qwen-1.5B}} \\ 
\midrule
DeepSeek-R1-Distill-Qwen-1.5B & 4252 & 33.0\% & 3346 & 45.9\% & 2388 & 45.3\% & 3314 & 33.0\% & 3456 & 43.6\% \\
AdaptThink-1.5B-delta0.05 & 4965 & 19.6\% & 3660 & 26.9\% & 3774 & 20.4\% & 3278 & \textbf{10.8\%} & 4419 & 30.7\% \\
L1-Qwen-1.5B-Max & 2553 & 15.4\% & 1955 & 31.4\% & 1605 & 36.7\% & 2422 & 26.8\% & 1977 & 28.8\% \\
TrainingEfficient\_alpha\_0.1\_DS-1.5B & 3452 & 14.9\% & 2315 & 24.5\% & 1433 & 25.6\% & 1924 & 18.1\% & 2477 & 26.6\% \\
DS-1.5B-thinkprune-iter2k & 3316 & \underline{13.2\%} & 2183 & \underline{19.8\%} & 1313 & 25.9\% & 1611 & 18.5\% & 2307 & \underline{23.6\%} \\
Laser-L8192-1.5B & 4750 & 22.5\% & 3474 & 33.3\% & 2167 & 33.5\% & 3122 & 23.9\% & 3481 & 31.9\% \\
Laser-DE-L4096-1.5B & 3767 & 16.2\% & 2485 & 22.8\% & 1442 & \underline{22.9\%} & 1785 & 15.8\% & 2606 & \underline{23.6\%} \\
DLER-R1-1.5B-Research & 2979 & 14.1\% & 2159 & 27.6\% & 1457 & 28.4\% & 1836 & 19.3\% & 2216 & 25.4\% \\
\textbf{APR-1.5B (Ours)} & 3207 & \textbf{10.5\%} & 2073 & \textbf{14.2\%} & 1184 & \textbf{15.6\%} & 1574 & \underline{14.3\%} & 2019 & \textbf{16.5\%} \\
\midrule

\rowcolor{gray!10} \multicolumn{11}{c}{\textit{Based on DeepSeek-R1-Distill-Qwen-7B}} \\ 
\midrule
DeepSeek-R1-Distill-Qwen-7B & 4103 & 28.0\% & 3285 & 42.5\% & 2384 & 45.3\% & 2990 & 34.1\% & 3236 & 42.0\% \\
Qwen3-8B & 5540 & 53.9\% & 4135 & 64.9\% & 3028 & 56.1\% & 3480 & 38.0\% & 3867 & 54.9\% \\
QwQ-32B & 5300 & 45.6\% & 4080 & 58.6\% & 2637 & 51.2\% & 3083 & 38.0\% & 3803 & 50.4\% \\
AdaptThink-7B-delta0.05 & 4008 & 20.7\% & 3200 & 33.0\% & 2761 & 34.3\% & 2407 & 24.3\% & 3274 & 31.0\% \\
L1-Qwen-7B-Max & 3284 & 12.8\% & 2248 & 28.9\% & 1736 & 38.7\% & 1723 & 24.5\% & 2332 & 28.2\% \\
TrainingEfficient\_alpha\_0.1\_DS-7B & 3859 & 20.4\% & 2693 & 33.7\% & 1741 & 34.9\% & 2162 & 25.5\% & 2848 & 33.0\% \\
SB\_DS7B\_alpha\_2 & 2990 & \textbf{8.4\%} & 1724 & \textbf{12.7\%} & 1090 & \textbf{13.1\%} & 840 & \textbf{7.0\%} & 1841 & \textbf{15.5\%} \\
Laser-DE-L4096-7B & 3528 & 13.7\% & 2168 & 22.6\% & 1247 & 23.8\% & 1456 & \underline{16.0\%} & 2341 & 22.9\% \\
DLER-R1-7B-Research & 2821 & 15.1\% & 1819 & 28.5\% & 1128 & 27.8\% & 1442 & 21.8\% & 1935 & 26.4\% \\
\textbf{APR-7B (Ours)} & 2606 & \underline{9.2\%} & 1816 & \underline{21.2\%} & 1144 & \underline{23.7\%} & 1254 & 17.1\% & 1793 & \underline{21.0\%} \\
\bottomrule
\end{tabular}}
    \label{tab:redundancy_ratio}
\end{table*}

\subsection{Comparison of Anchor Localization Methods}
\label{app:localization}
Fig.~\ref{fig:localization} (a) empolys DS-1.5B trained with DAPO for 250 steps, with the length-penalty coefficient $\beta = 2\text{e-}4$; Fig.~\ref{fig:localization} (b) empolys DS-7B trained with DAPO for 100 steps, with the length-penalty coefficient $\beta = 5\text{e-}4$.
\definecolor{qred}{HTML}{F57C6E}
\definecolor{qgreen}{HTML}{84C3B7}
\definecolor{qblue}{HTML}{71B8ED}
\definecolor{qpink}{HTML}{F2A8DA}

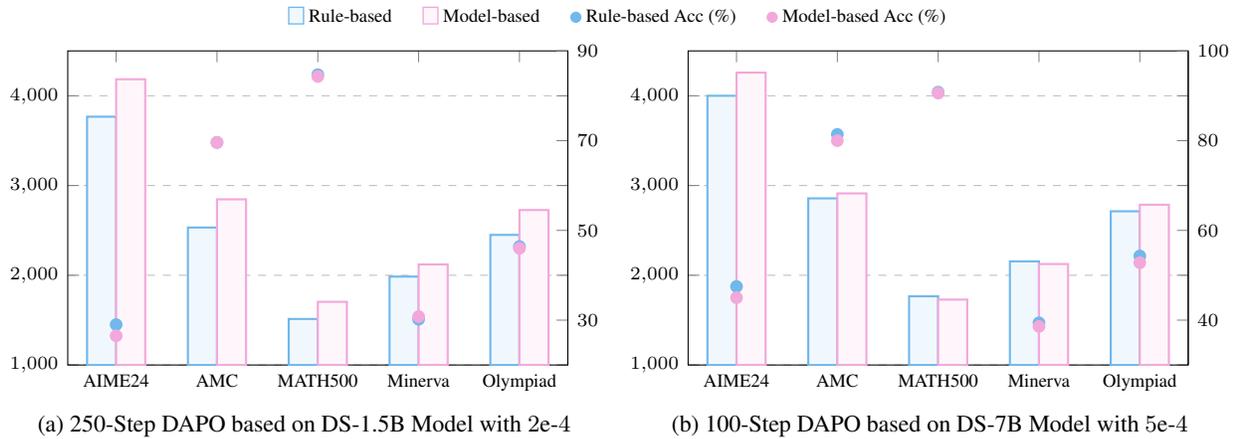
\begin{figure}
\centering

\begin{minipage}{.6\textwidth}
\centering
\begin{tikzpicture}[baseline]
\node[font=\scriptsize] {
    \raisebox{-0.3ex}{\tikz \draw[fill=qblue!10, draw=qblue, line width=0.5pt] (0,0) rectangle (0.8em,2ex);} Rule-based
    \hspace{8pt}
    \raisebox{-0.3ex}{\tikz \draw[fill=qpink!10, draw=qpink, line width=0.5pt] (0,0) rectangle (0.8em,2ex);} Model-based
    \hspace{8pt}
\tikz[baseline=-0.5ex] \node[circle, fill=qblue, inner sep=1.5pt] {}; Rule-based Acc (\%)
    \hspace{8pt}
\tikz[baseline=-0.5ex] \node[circle, fill=qpink, inner sep=1.5pt] {}; Model-based Acc (\%)
};
\end{tikzpicture}
\end{minipage}

\begin{minipage}{.48\textwidth}
\centering
\begin{tikzpicture}
\begin{axis}[
    ybar,
    bar width=11pt,
    width=\textwidth,
    height=.7\textwidth,
    ymajorgrids,
    grid style=dashed,
    symbolic x coords={AIME24,AMC,MATH500,Minerva,Olympiad},
    xtick=data,
    tick align=inside,
    y tick label style={font=\scriptsize},
    x tick label style={rotate=0, anchor=north, font=\scriptsize},
    enlarge x limits=0.12,
    ymin=1000, ymax=4500,
    ylabel style={font=\scriptsize},
]

\addplot[fill=qblue!10, draw=qblue, bar shift=-5.5pt, line width=0.8pt] 
coordinates {(AIME24,3767) (AMC,2532) (MATH500,1513) (Minerva,1985) (Olympiad,2450)};

\addplot[fill=qpink!10, draw=qpink, bar shift=5.5pt, line width=0.8pt] 
coordinates {(AIME24,4184) (AMC,2846) (MATH500,1704) (Minerva,2121) (Olympiad,2727)};

\end{axis}

\begin{axis}[
    axis y line*=right,
    axis x line=none,
    ymin=20, ymax=90,
    ytick={30,50,70,90},
    ylabel style={font=\scriptsize},
    yticklabel style={font=\scriptsize},
    enlarge x limits=0.12,
    symbolic x coords={AIME24,AMC,MATH500,Minerva,Olympiad},
    grid=none,
    width=\textwidth,
    height=.7\textwidth,
]

\addplot[color=qblue, mark=*, mark size=2pt, line width=1pt, thick,draw=none]
coordinates {(AIME24,29.0) (AMC,69.6) (MATH500,84.7) (Minerva,30.2) (Olympiad,46.4)};

\addplot[color=qpink, mark=*, mark size=2pt, line width=1pt, thick,draw=none]
coordinates {(AIME24,26.5) (AMC,69.6) (MATH500,84.3) (Minerva,30.8) (Olympiad,46.0)};

\end{axis}
\end{tikzpicture}
{\small \quad (a) 250-Step DAPO based on DS-1.5B Model with $2\text{e-}4$}
\end{minipage}
\begin{minipage}{.48\textwidth}
\centering
\begin{tikzpicture}
\begin{axis}[
    ybar,
    bar width=11pt,
    width=\textwidth,
    height=.7\textwidth,
    ymajorgrids,
    grid style=dashed,
    symbolic x coords={AIME24,AMC,MATH500,Minerva,Olympiad},
    xtick=data,
    tick align=inside,
    y tick label style={font=\scriptsize},
    x tick label style={rotate=0, anchor=north, font=\scriptsize},
    enlarge x limits=0.12,
    ymin=1000, ymax=4500,
    ylabel style={font=\scriptsize},
]

\addplot[fill=qblue!10, draw=qblue, bar shift=-5.5pt, line width=0.8pt] 
coordinates {(AIME24,4000) (AMC,2857) (MATH500,1766) (Minerva,2155) (Olympiad,2713)};

\addplot[fill=qpink!10, draw=qpink, bar shift=5.5pt, line width=0.8pt] 
coordinates {(AIME24,4258) (AMC,2912) (MATH500,1729) (Minerva,2125) (Olympiad,2786)};

\end{axis}

\begin{axis}[
    axis y line*=right,
    axis x line=none,
    ymin=30, ymax=100,
    ylabel style={font=\scriptsize},
    yticklabel style={font=\scriptsize},
    enlarge x limits=0.12,
    symbolic x coords={AIME24,AMC,MATH500,Minerva,Olympiad},
    grid=none,
    width=\textwidth,
    height=0.7\textwidth,
]

\addplot[color=qblue, mark=*, mark size=2pt, line width=1pt, thick,draw=none]
coordinates {(AIME24,47.5) (AMC,81.4) (MATH500,90.8) (Minerva,39.4) (Olympiad,54.3)};

\addplot[color=qpink, mark=*, mark size=2pt, line width=1pt, draw=none,thick]
coordinates {(AIME24,45) (AMC,80) (MATH500,90.6) (Minerva,38.6) (Olympiad,52.8)};

\end{axis}
\end{tikzpicture}
{\small \quad (b) 100-Step DAPO based on DS-7B Model with $5\text{e-}4$}
\end{minipage}

\caption{Comparison of Rule-based and Model-based Anchor Localization Methods (Left y-axis: average generation length; Right y-axis: average Pass@1 accuracy, averaged over 16 samples).}
\label{fig:localization}
\end{figure}

\section{Implementation Details of Anchor Localization}
\subsection{Keywords for Rule-based Localization}
\label{app:keywords}
\begin{tcolorbox}[colback=gray!5!white,colframe=gray!75!black]
\textbf{Conclusion Keywords:} \\
`\texttt{therefore}', `\texttt{thus}', `\texttt{hence}', `\texttt{so}', `\texttt{answer}', `\texttt{solution}', `\texttt{result}', `\texttt{final}', `\texttt{indeed}', \\
`\texttt{conclude}', `\texttt{equals}', `\texttt{valid}', `\texttt{set}', `\texttt{maybe}', `\texttt{seem}', `\texttt{perhaps}', `\texttt{we get}', `\texttt{we have}', \\
`\texttt{i get}', `\texttt{would be}', `\texttt{should be}', `\texttt{it is}', `\texttt{it's}', `\texttt{that's}', `\texttt{lead to}', `\texttt{value of}', \\
`\texttt{the only}', `\texttt{correct option}', `\texttt{maximum possible}'
\vspace{\baselineskip} \\
\textbf{Verification Keywords:} \\
`\texttt{check}', `\texttt{verify}', `\texttt{confirm}', `\texttt{wait}', `\texttt{make sure}', `\texttt{double-check}', `\texttt{let me}', `\texttt{let's}', \\
 `\texttt{straightforward}', `\texttt{miss anything}', `\texttt{is that right}', `\texttt{is that correct}', `\texttt{is that all}'

\end{tcolorbox}

\subsection{Training Details for Anchor Locator}
\label{app:locator}
\paragraph{Data Construction for Anchor Locator.}
We synthesized the SFT dataset for the Anchor Locator through a rigorous two-stage filtering process.

\textbf{Stage 1: Initial Sampling and Screening.} We first randomly sampled 30,000 queries from the DeepScaleR-Preview dataset and divided them equally into three subsets. For each subset, we generated a single reasoning response using DS-1.5B, DS-7B, and Qwen3-8B, respectively. From these raw outputs, we filtered out responses missing the closing \texttt{</think>} tag and selected 5,000 valid reasoning traces from each model (totaling 15,000 samples) as the input for annotation.

We then prompted the advanced LLM Gemini3-Flash\footnote{\url{https://deepmind.google/models/gemini/flash}} with the temperature of 1 to identify and extract the specific sentence where the final answer first emerges within the reasoning trace (see Appendix~\ref{app:prompt} for the full prompt template).

\textbf{Stage 2: Consistency Filtering.} Upon obtaining the extractions from Gemini3-Flash, we performed a strict validity check to filter out instances where the extracted sentence did not appear verbatim in the original reasoning trace (indicating hallucination). This process resulted in a final high-quality dataset comprising 12,000 training samples and 500 held-out test samples. 

We utilize the Swift framework to fine-tune Qwen3-8B. The hyperparameters used are detailed in Table~\ref{tab:sft_params}. The model input is constructed by concatenating the the same prompt template used for data construction with the problem statement.

\begin{table*}[h]
    \centering
    \caption{Training hyperparameters for supervised fine-tuning Qwen3-8B.}
    \resizebox{0.4\textwidth}{!}{%
    \begin{tabular}{l c}

\toprule
\textbf{Parameter} & \textbf{Value}\\
\midrule
\texttt{model\_type} & qwen3\\
\texttt{train\_type} & full\\
\midrule
\texttt{torch\_dtype} & bfloat16\\
\midrule
\texttt{split\_dataset\_ratio} & 0\\
\texttt{max\_length} & 10000\\
\midrule
\texttt{num\_train\_epochs} & 2\\
\texttt{per\_device\_train\_batch\_size} & 4\\
\texttt{per\_device\_eval\_batch\_size} & 4\\
\texttt{learning\_rate} & 2e-5\\
\texttt{gradient\_accumulation\_steps} & 1\\
\texttt{warmup\_ratio} & 0.01\\
\texttt{seed} & 42\\
\midrule
\texttt{eval\_strategy} & steps\\
\texttt{eval\_steps} & 0.1\\
\bottomrule

\end{tabular}%
}
    \label{tab:sft_params}
\end{table*}

\section{Policy Optimization Algorithms}
\label{app:algo}
\paragraph{Group Relative Policy Optimization.}
GRPO~\cite{shao2024deepseekmath} is the currently mainstream method, estimating advantage via intra-group normalization.
GRPO estimates advantages in a group-relative manner and avoids fitting an explicit value function.
For each query--answer pair $(q,a)$, the behavior policy $\pi_{\theta_{\text{old}}}$ samples a rollout group of $G$ responses
$\{o_i\}_{i=1}^G \sim \pi_{\theta_{\text{old}}}(\cdot\mid q)$.
Let $R_i$ denote the reward assigned to $o_i$.
The advantage for the $i$-th response is obtained by normalizing rewards within the group:
\begin{equation}
\hat A_i
=
\frac{
R_i - \mathrm{mean}\!\left(\{R_i\}_{i=1}^{G}\right)
}{
\mathrm{std}\!\left(\{R_i\}_{i=1}^{G}\right)
}.
\label{eq:grpo-adv}
\end{equation}
GRPO updates the policy by maximizing the following objective:
\begin{equation}
\mathcal{J}_{\mathrm{GRPO}}(\theta)
=\mathbb{E}_{(q,a)\sim\mathcal{D},\{o_i\}_{i=1}^G\sim \pi_{\theta_{\text{old}}}(\cdot\mid q)}
\Bigg[
\frac{1}{G}\sum_{i=1}^G \frac{1}{|o_i|}
\sum_{t=1}^{|o_i|} 
\min\!\Big(
r_{i,t}(\theta)\hat A_i,\
\mathrm{clip}\!\big(r_{i,t}(\theta),1-\epsilon,1+\epsilon\big)\hat A_i
\Big)
\Bigg],
\label{eq:grpo-obj}
\end{equation}
where \(\epsilon\) is the clipping range of importance sampling ratio:
\begin{equation}
r_{i,t}(\theta)
=
\frac{\pi_\theta(o_{i,t}\mid q, o_{i,<t})}{\pi_{\theta_{\text{old}}}(o_{i,t}\mid q, o_{i,<t})}.
\label{eq:grpo-ratio}
\end{equation}

\paragraph{Direct Alignment Policy Optimization.}
DAPO~\cite{yu2025dapo} augments group-based policy optimization with a dynamic sampling strategy to avoid uninformative batches. 
Concretely, it repeatedly samples prompts and discards groups whose within-group accuracy is degenerate (i.e., all correct or all incorrect), retaining only those with mixed outcomes so that each update batch contains effective learning signals. 
As a result, the number of sampling attempts per batch becomes adaptive: sampling continues until the batch is filled with prompts whose group accuracy is neither 0 nor 1.

\begin{equation}
\begin{aligned}
\mathcal{J}_{\mathrm{DAPO}}(\theta) =\quad& \mathbb{E}_{(q,a)\sim \mathcal{D}, \{o_i\}_{i=1}^G\sim \pi_{\theta_\text{old}}(\cdot\mid q)}
\Bigg[\frac{1}{\sum_{i=1}^{G}|o_i|}\sum_{i=1}^{G}\sum_{t=1}^{|o_i|} 
\min \Big( r_{i,t}(\theta) \hat{A}_{i,t},  
\ \text{clip} \Big( r_{i,t}(\theta), 1 - {\varepsilon_{\text{low}}}, 1 + {\varepsilon_{\text{high}}} \Big) \hat{A}_{i,t} \Big) \Bigg]
\\
\text{s.t.}\quad& {\color{red}0< \Big|\{o_i\mid\texttt{is\_equivalent}(a,o_i)\}\Big|< G}.
\label{eq:dapoloss_oversample_filter}
\end{aligned}
\end{equation}
where
\begin{equation}
    r_{i,t}(\theta)=\frac{\pi_{\theta}(o_{i,t} \mid q, o_{i,<t})}{\pi_{\theta_{\text{old}}}(o_{i,t} \mid q,o_{i,<t})},\quad\hat{A}_{i,t} = \frac{R_i - \text{mean}(\{R_i\}_{i=1}^G)}{\text{std}(\{R_i\}_{i=1}^G)}.
\label{eq:advantage_calculation}
\end{equation}

\section{Experimental Details}
\subsection{Detailed Formulation for Accuracy-Efficiency (AE) Score}
\label{app:ae_score}

\paragraph{Accuracy-Efficiency (AE) Score.}
To evaluate the trade-off between inference efficiency and task performance, we adopt the Accuracy-Efficiency (AE) Score \cite{luo2025o1}. This metric serves as a unified measure to determine whether a model effectively reduces generation length without sacrificing reasoning accuracy. The scoring function is formulated as:

\begin{equation}
\label{eq:ae_score}
\text{AE Score} = 
\begin{cases} 
\varphi \cdot \Delta \text{Length} + \eta \cdot |\Delta \text{Acc}|, & \text{if } \Delta \text{Acc} \geq 0 \\
\varphi \cdot \Delta \text{Length} - \theta \cdot |\Delta \text{Acc}|, & \text{if } \Delta \text{Acc} < 0 
\end{cases}
\end{equation}
where $\Delta \text{Length}$ and $\Delta \text{Acc}$ represent the percentage changes in output length and accuracy relative to the base model, respectively. 
We adhere to the hyperparameter settings recommended by \citet{luo2025o1}: $\varphi = 1$ (weight for length reduction), $\eta = 3$ (bonus for accuracy gains), and $\theta = 5$ (penalty for accuracy drops). 
The asymmetric design, where $\theta > \eta$, ensures that the metric imposes a heavier penalty on performance regression than gains, thereby prioritizing reasoning performance.

\subsection{Computational Resources}
To accommodate the computational demands of different anchor localization strategies, experiments with Rule-based localization are conducted on eight H100 (80GB) GPUs, while those with Model-based localization are trained on eight H200 (141GB) GPUs.

\subsection{Detailed Descriptions of Baselines}
\label{app:baselines}
\paragraph{Baseline Descriptions.}
The baselines employed in our experiments are characterized as follows:

\begin{itemize}[leftmargin=*]
    \item \textbf{DeepSeek-R1-Distill-Qwen-1.5/7B}: The base models without further post-training, serving as the reference for initial performance.
    
    \item \textbf{AdaptThink-1.5/7B-delta0.05}: AdaptThink~\cite{zhang2025adaptthink} encourages the model to adaptively adjust its generation length according to the difficulty of the problem.
    
    \item \textbf{L1-Qwen-1.5/7B-Max}: The best-performing model reported in L1~\cite{aggarwal2025l1}, which utilizes prompt design to control the computational budget. The model is initialized from DeepScaleR-1.5B-Preview and trained with a context window of 4,096 tokens.
    
    \item \textbf{TrainingEfficient\_alpha\_0.1\_DS-1.5/7B}: TrainingEfficient~\cite{arora2025training} adopts the group-wise normalized length as the reference and utilizes PPO with the RLOO advantage estimator for RL training.
    
    \item \textbf{DS-1.5B-thinkprune-iter2k}: ThinkPrune~\cite{hou2025thinkprune} iteratively tightens the length budget across training rounds, reducing the limit from 4k to 3k, and finally to 2k tokens.
    
    \item \textbf{SB\_DS7B\_alpha\_2}: ShorterBetter~\cite{yi2025shorterbetter} selects the length of the shortest correct response within a group as the reference. It is trained with a context window of 6k tokens.
    
    \item \textbf{Laser-L8192-1.5B, Laser-DE-L4096-1.5/7B}: Laser~\cite{liu2025learnreasonefficientlyadaptive} introduces a difficulty-aware length penalty reward. Note that the former configuration (L8192) does not have a released 7B version.
    
    \item \textbf{DLER-R1-1.5/7B-Research}: DLER~\cite{liu2025dler} optimizes the RL algorithm using a simple truncation length penalty (set to 2k/4k tokens).
\end{itemize}

\subsection{Prompt Template}
\label{app:prompt}
Following the principles of efficient prompting outlined in~\citet{chang2024efficient}, we designed prompts for LLMs to extract the sentence where the reference answer first appears in the reasoning trace.

\begin{tcolorbox}[width=\textwidth, colback=black!5!white, colframe=black!75!black, title=Prompt for extracting the first sentence containing the final answer.]
\noindent
You are a reasoning trace analyst. Your role is to identify the first sentence where the model gets the answer of a problem. The goal is to identify the redundant overthinking process after the model has actually solved the problem. \\

You will be given a reasoning trace, which ends with the `\texttt{</think>}' tag; you will also be given the final answer. You must: \\

1. Identify only the \textbf{first} sentence where the model gets the final answer.

2. The sentence you return should be \textbf{exactly the same} as the one in the original reasoning trace.

3. If the sentence containing the final answer is not found, please return \textbf{NULL}. \\

Return only the sentence you identify, no extra commentary or explanation. \\

Final Answer:

\{final\_answer\} \\

Reasoning trace:

\{thinking\}
\end{tcolorbox}

\begin{tcolorbox}[width=\textwidth, colback=black!5!white, colframe=black!75!black, title=Prompt for extracting the first sentence containing a complete answer.]
\noindent
You are a reasoning trace analyst. Your role is to identify the first sentence where the model gets the answer of a problem. The goal is to identify the redundant overthinking process after the model has actually solved the problem. \\

You will be given a reasoning trace, which ends with the `\texttt{</think>}' tag. You must: \\

1. Identify only the \textbf{first} sentence where the model gets the answer of the problem.

2. The sentence you return should be \textbf{exactly the same} as the one in the original reasoning trace. \\

Return only the sentence you identify, no extra commentary or explanation. \\

Reasoning trace:

\{thinking\}
\end{tcolorbox}

\begin{tcolorbox}[width=\textwidth, colback=black!5!white, colframe=black!75!black, title=Prompt for LRM reasoning during training and evaluation.]
\noindent
Please reason step by step, and put your final answer within \textbackslash boxed\{\}.
\end{tcolorbox}

\subsection{Hyperparameter Settings}
\label{app:hyperparameters}

\begin{table*}[h]
    \centering
    \caption{Training hyperparameters for training our APR models based on RL.}
    \resizebox{0.7\textwidth}{!}{%
    \begin{tabular}{l c c}

\toprule
\textbf{Parameter} & \textbf{GRPO} & \textbf{DAPO}\\
\midrule
\texttt{algorithm.adv\_estimator} & grpo & grpo\\
\texttt{actor\_rollout\_ref.actor.loss\_agg\_mode} & token-mean & token-mean\\
\texttt{actor\_rollout\_ref.actor.use\_kl\_loss} & True & False\\
\texttt{actor\_rollout\_ref.actor.kl\_loss\_type} & low\_var\_kl & low\_var\_kl\\
\texttt{actor\_rollout\_ref.actor.kl\_loss\_coef} & 0.001 & 0.001\\
\texttt{actor\_rollout\_ref.actor.entropy\_coeff} & 0 & 0\\
\texttt{actor\_rollout\_ref.actor.grad\_clip} & 1.0 & 1.0\\
\texttt{actor\_rollout\_ref.actor.clip\_ratio\_low} & 0.2 & 0.2\\
\texttt{actor\_rollout\_ref.actor.clip\_ratio\_high} & 0.2 & 0.28\\
\texttt{actor\_rollout\_ref.actor.clip\_ratio\_c} & 3.0 & 10.0\\
\midrule
\texttt{actor\_rollout\_ref.actor.optim.lr} & 1e-6 & 1e-6\\
\texttt{actor\_rollout\_ref.actor.optim.lr\_warmup\_steps} & -1 & 10\\
\texttt{actor\_rollout\_ref.actor.optim.weight\_decay} & 0.01 & 0.1\\
\midrule
\texttt{algorithm.use\_kl\_in\_reward} & True & False\\
\texttt{algorithm.kl\_ctrl.kl\_coef} & 0.001 & 0\\
\texttt{algorithm.filter\_groups.enable} & - & True\\
\texttt{algorithm.filter\_groups.max\_num\_gen\_batches} & - & 10\\
\texttt{algorithm.filter\_groups.metric} & - & acc\\
\midrule
\texttt{data.train\_batch\_size} & 128 & 128\\
\texttt{data.val\_batch\_size} & 512 & -\\
\texttt{data.gen\_batch\_size} & - & 384\\
\texttt{actor\_rollout\_ref.actor.ppo\_mini\_batch\_size} & 64 & 64\\
\texttt{actor\_rollout\_ref.actor.ppo\_epochs} & 1 & 1\\
\midrule
\texttt{data.max\_prompt\_length} & 2048 & 2048\\
\texttt{data.max\_response\_length} & 8192 & 8192\\
\midrule
\texttt{actor\_rollout\_ref.rollout.n} & 8 & 8\\
\texttt{actor\_rollout\_ref.rollout.temperature} & 0.9 & 0.9\\
\texttt{actor\_rollout\_ref.rollout.top\_p} & 1.0 & 1.0\\
\texttt{actor\_rollout\_ref.rollout.top\_k} & -1 & -1\\
\midrule
\texttt{reward\_model.reward\_kwargs.beta} & 2e-4/5e-4 & 2e-4/5e-4\\
\bottomrule

\end{tabular}%
}
    \label{tab:rl_params}
\end{table*}

\begin{table*}[h]
    \centering
    \caption{Training hyperparameters for training L1-Qwen-1.5B-Max based on RL.}
    \resizebox{0.7\textwidth}{!}{%
    \begin{tabular}{l c c}

\toprule
\textbf{Parameter} & \textbf{L1\_Exact} & \textbf{L1\_Max}\\
\midrule
\texttt{algorithm.adv\_estimator} & grpo & grpo\\
\texttt{actor\_rollout\_ref.actor.use\_kl\_loss} & True & True\\
\texttt{actor\_rollout\_ref.actor.kl\_loss\_type} & low\_var\_kl & low\_var\_kl\\
\texttt{actor\_rollout\_ref.actor.kl\_loss\_coef} & 0.001 & 0.001\\
\texttt{actor\_rollout\_ref.actor.optim.lr} & 1e-6 & 4e-6\\
\midrule
\texttt{actor\_rollout\_ref.model.use\_remove\_padding} & True & False\\
\midrule
\texttt{trainer.total\_epochs} & 3 & 1\\
\midrule
\texttt{data.train\_batch\_size} & 128 & 128\\
\texttt{data.val\_batch\_size} & 512 & 512\\
\texttt{actor\_rollout\_ref.actor.ppo\_mini\_batch\_size} & 64 & 64\\
\midrule
\texttt{data.max\_prompt\_length} & 1024 & 1024\\
\texttt{data.max\_response\_length} & 4096 & 4500\\
\midrule
\texttt{actor\_rollout\_ref.rollout.n} & 16 & 16\\
\texttt{actor\_rollout\_ref.rollout.temperature} & 0.6 & 0.6\\
\texttt{actor\_rollout\_ref.rollout.enforce\_eager} & False & True\\
\texttt{actor\_rollout\_ref.rollout.free\_cache\_engine} & False & True\\
\midrule
\texttt{reward\_config.sigmoid\_reward} & False & -\\
\texttt{reward\_config.linear\_reward} & True & -\\
\texttt{reward\_config.multiplier\_reward} & False & -\\
\texttt{reward\_config.alpha} & 0.0003 & -\\
\bottomrule

\end{tabular}%
}
    \label{tab:l1_params}
\end{table*}


\end{document}